\documentclass[journal]{IEEEtran}
\usepackage[utf8]{inputenc}

\usepackage[english]{babel}
\usepackage{amsmath, amssymb}
\usepackage{float}
\usepackage{makecell}
\usepackage{multirow}
\usepackage{cite}
\usepackage{amsmath,amssymb,amsfonts}
\usepackage{algorithmicx}
\usepackage{amsmath}
\usepackage{amsfonts}
\usepackage{amssymb}\usepackage{graphicx}
\usepackage{textcomp}
\usepackage{xcolor}

\usepackage{subfigure}
\usepackage{epstopdf}
\usepackage{aecompl}
\usepackage{times}
\usepackage{epsfig}
\usepackage{graphicx}
\usepackage[T1]{fontenc}    
\usepackage{url}            
\usepackage{booktabs}       
\usepackage{nicefrac}       
\usepackage[ruled,vlined]{algorithm2e}

\ifCLASSINFOpdf
 
\else
 
\fi

\begin{document}
%


\title{Anchor-Based Spatio-Temporal Attention 3D Convolutional Networks for Dynamic 3D Point Cloud Sequences}

\author{Guangming Wang, Muyao Chen, Hanwen Liu, Yehui Yang, Zhe Liu, and Hesheng Wang
	\thanks{This work was supported in part by the Natural Science Foundation of China under Grant U1613218, 61722309, and U1913204; in part by Beijing Advanced Innovation Center for Intelligent Robots and Systems under Grant 2019IRS01; and in part by grants from the NVIDIA Corporation. Corresponding Author: Hesheng Wang (e-mail: wanghesheng@sjtu.edu.cn).}
	\thanks{G. Wang, Y. Yang, and H. Wang are with the Department of Automation, Institute of Medical Robotics, Key Laboratory of System Control and Information Processing of Ministry of Education, Key Laboratory of Marine Intelligent Equipment, and System of Ministry of Education, Shanghai Jiao Tong University, Shanghai 200240, China. H. Wang is with Beijing Advanced Innovation Center for Intelligent Robots and Systems, Beijing Institute of Technology, China. M. Chen and H. Liu are with the Department of Computer Science and Engineering, Shanghai Jiao Tong University, Shanghai 200240, China.
		
	Z. Liu is with the Department of Computer Science and Technology, University of Cambridge.}}

\markboth{Journal of \LaTeX\ Class Files,~Vol.~14, No.~8, August~2015}%
{Shell \MakeLowercase{\textit{et al.}}: Bare Demo of IEEEtran.cls for IEEE Journals}

\maketitle

\begin{abstract}
    With the rapid development of measurement technology,  LiDAR and depth cameras are widely used in the perception of the 3D environment. Recent learning based methods for robot perception most focus on the image or video, but deep learning methods for dynamic 3D point cloud sequences are underexplored. Therefore, developing efficient and accurate perception method compatible with these advanced instruments is pivotal to autonomous driving and service robots.
	An Anchor-based Spatio-Temporal Attention 3D Convolution operation (ASTA3DConv) is proposed in this paper to process dynamic 3D point cloud sequences. The proposed convolution operation builds a regular receptive field around each point by setting several virtual anchors around each point. The features of neighborhood points are firstly aggregated to each anchor based on the spatio-temporal attention mechanism. Then, anchor-based 3D convolution is adopted to aggregate these anchors' features to the core points. The proposed method makes better use of the structured information within the local region and learns spatio-temporal embedding features from dynamic 3D point cloud sequences. Anchor-based Spatio-Temporal Attention 3D Convolutional Neural Networks (ASTA3DCNNs) are built for classification and segmentation tasks based on the proposed ASTA3DConv and evaluated on action recognition and semantic segmentation tasks. The experiments and ablation studies on MSRAction3D and Synthia datasets demonstrate the superior performance and effectiveness of our method for dynamic 3D point cloud sequences. Our method achieves the state-of-the-art performance among the methods with dynamic 3D point cloud sequences as input on MSRAction3D and Synthia datasets.
\end{abstract}


%
\IEEEpeerreviewmaketitle

\section{Introduction}
The measurement and understanding of the 3D environment, such as action recognization \cite{zhang2019knowledge,chen2019smartphone} and semantic segmentation \cite{qiu2018rgb}, are essential for autonomous driving and service robots. Some mainstream sensors, such as depth cameras and LiDAR, can directly get the dynamic 3D point cloud sequences of the surrounding environment, containing much spatio-temporal information. Thus, studying effective feature extraction and measurement methods for the 3D point clouds has been the research community's focus in recent years. The latest works \cite{qi2017pointnet,qi2017pointnet++} show the potential of directly consuming points, and they do not need to convert the point clouds into other forms, such as voxel form\cite{wang2017cnn,riegler2017octnet,graham20183d}. Many works have explored the learning of single point cloud on 3D object retrieval~\cite{kuang2019effective}, classification~\cite{qi2017pointnet,qi2017pointnet++}, and segmentation~\cite{mao2019interpolated,hu2019randla,wang2020spherical}. There are also a few research works on the learning of multi-frame point cloud \cite{choy20194d,luo2018fast,liu2019meteornet}, but there remain some challenges.

Some methods \cite{choy20194d,luo2018fast}  convert point clouds into grid representations. The grid quantization error is inevitable. Moreover, the extra conversion will cause inefficient processing performance. 
Latest work, MeteorNet\cite{liu2019meteornet} handles the point cloud sequences directly by adding time encoding with the position and feature encoding in PointNet++~\cite{qi2017pointnet++}. In MeteorNet~\cite{liu2019meteornet}, the chained-flow grouping relies on the accuracy of scene flow estimation, and the direct grouping uses varying radius for different frames. However, shared Multi-layer Perceptrons (MLP) and max pooling are still used, which lacks the ability to describe structurally and loses information \cite{mao2019interpolated,wang2020spherical,hu2019randla,wang2020hierarchical}. It is still challenging to extract structured features from irregular point cloud sequences without converting the data forms. This paper focuses on a novel 3D convolution method for raw 3D point cloud sequences. Inspired by the recent interpolated convolution methods~\cite{mao2019interpolated,wang2020spherical} on a single frame of 3D point clouds, an anchor-based 3D convolution is designed to deal with dynamic 3D point cloud sequences. 

Although spatio-temporal attention models have been widely used for image sequence based tasks, including person re-identification \cite{zhang2019multi}, video action recognition \cite{li2018unified}, video saliency detection \cite{zhong2013video}, and so on, there are still challenges for the spatio-temporal attention for point cloud sequences. The spatio-temporal attention in the video sequence does not need to consider specific points with different spatial distribution, while the unstructured 3D point cloud requires different attention to different spatial positions because these points are not uniformly distributed in 3D space. The previous multi-frame work MeteorNet~\cite{liu2019meteornet} only use max pooling for the feature aggregation after grouping, which losses information. We expect to make full use of information in an adaptive weighted method. Therefore, a new spatio-temporal attention model based on point cloud sequence is proposed in this paper.

In this paper, an anchor-based spatio-temporal attention 3D convolution (ASTA3DConv) is proposed to extract structural features by the spatio-temporal attentive embedding and customized 3D convolution form from the sparse and irregular point cloud sequence data. Specifically, multiple virtual anchors are established around each real point with the learnable convolution weights. The structural features from 3D points around these anchors are firstly learned and embedded into these virtual anchors by a spatio-temporal attentive embedding. The proposed spatio-temporal embedding method encodes spatial coordinates, timestamps, and point features to anchors' structured features. Then, a predefined 3D convolution based on the structure of anchors is adopted to extract structural features from these anchors to obtain the final features of each real point, the kernel center of the anchor-based convolution kernel. Compared with classic $K$ Nearest Neighbors (KNN) and ball query \cite{qi2017pointnet++}, the spatio-temporal attentive embedding structurally organizes the raw unordered and irregular data in a learnable fashion, and then the anchor-based 3D convolution method contributes to the 3D structure feature learning from irregular raw data. The proposed method makes it possible to learn 3D feature from irregular dynamic 3D point cloud sequences in a full learnable fashion, without the interpolation calculation~\cite{mao2019interpolated,wang2020spherical}.

Based on the ASTA3DConv, we propose Anchor-based Spatio-Temporal Attention 3D Convolutional Neural Networks (ASTA3DCNNs), including the classification network and segmentation network. 
The classification network can obtain the probabilities that the point cloud sequences belong to each category. The segmentation network can output the probabilities that each point in point cloud sequences belongs to each category. The category with the highest probability of each point serves as the segmentation category for each point. 
The contributions are as follows:
\begin{itemize}
	\item  A novel Anchor-based Spatio-Temporal Attention 3D Convolution operation (ASTA3DConv) is proposed in this paper. By introducing a discrete 3D convolution kernel on multiple virtual structured anchors after the feature aggregation into anchors, the 3D features can be structurally extracted from raw disordered 3D data without discretizing the data. 	
	\item To gather sparse, irregular, and unordered points to the virtual anchors, the spatio-temporal attentive embedding is proposed to learn the spatio-temporal embedding information aggregated to each anchor. The spatio-temporal attentive embedding considers the euclidean space, feature space, and time space. The soft weighted method is used to replace the classical max pooling \cite{liu2019meteornet} to make full use of information.
	\item The Anchor-based Spatio-Temporal Attention 3D Convolutional Neural Networks (ASTA3DCNNs) are further proposed for the action recognization and semantic segmentation tasks. Experiments on MSRAction3D dataset\cite{li2010action} and Synthia dataset\cite{ros2016synthia} show that the proposed networks achieves superior performance compared with the state-of-the-art methods. The ablation studies demonstrate the effectiveness of each design.
\end{itemize}

\begin{figure*}[t]
	\centering
	\vspace{0mm}
	\resizebox{0.90\textwidth}{!}
	{
		\includegraphics[scale=1.00]{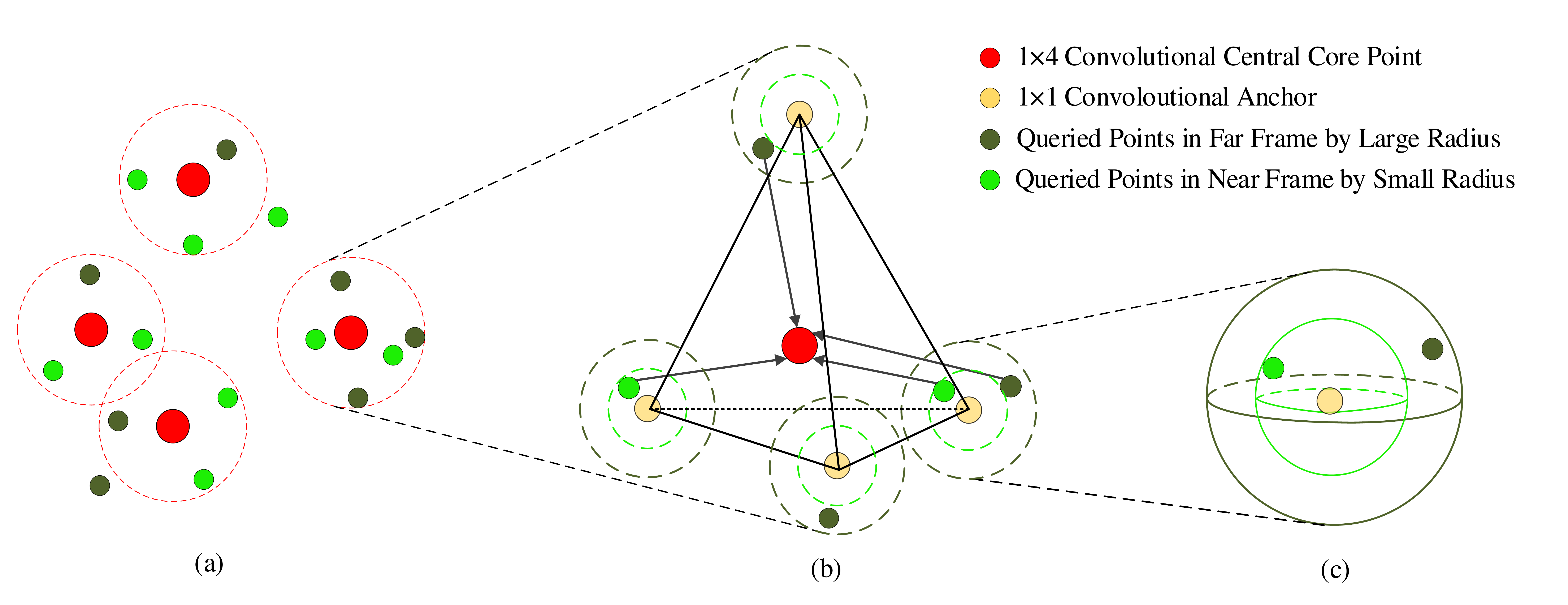}}
	\vspace{-4mm}
	\caption{The overview of the proposed anchor-based spatio-temporal attention convolutional operation for dynamic 3D point cloud sequences. (a) Input 3D point clouds from multiple frames. Red points represent the selected core points. Light green and dark green points represent points from the near and far frame. (b) The details of constructing virtual anchors. Light yellow points represent the 4 anchors constructed as a regular tetrahedron form. (c) Querying neighborhood points around an anchor from multiple frames. As shown in Section \ref{section:2}, different radii are applied for the ball querying from different frames.
}
	\vspace{-2mm}
	\label{fig:anc3}
\end{figure*}

\section{Related Work}\label{related}

\subsection{Deep Learning on Single Frame of Point Clouds}
PointNet~\cite{qi2017pointnet} is a pioneering work in applying deep learning to consume point clouds directly. The primary approach is to construct a symmetric function. It uses shared MLP to aggregate each point's neighborhood information and then uses an element-wise max pooling to extract local features. The continuation work PointNet++~\cite{qi2017pointnet++} extracted and aggregated neighborhood features hierarchically in the Euclidean space around each point. 

Recent works made some innovations in the design of the convolution kernel. SPLATNet~\cite{su2018splatnet} used the high-dimension mesh to carry features of input points and then extracted features by adopting the bilateral convolution.
SpiderCNN~\cite{xu2018spidercnn} proposed to apply different weights to each convolution kernel for each neighbor.
Komarichev et al.~\cite{komarichev2019cnn} utilized annular convolution for point clouds before regular convolution operations by considering orbicular structures and directions. Method of Lei et al.~\cite{lei2019octree} used spherical convolution kernel to separate the space and extracted features. Zhao et al.~\cite{zhao2019pointweb} applied the dense connection between every two points in the local neighborhood by Adaptive Feature Adjustment (AFA) to represent better than DGCNN~\cite{wang2019dynamic}. ShellNet~\cite{zhang2019shellnet} used the max pooling for overall points in each concentric spherical shell.
 Unlike these works, ours focuses on shared discrete convolution on the feature extraction of raw 3D point clouds, like the successful convolutional neural networks (CNNs) on 2D images. However, different from the voxel-based methods \cite{su2018splatnet,lei2019octree,liu2019pvcnn,xie2020grnet}, our method can directly consume raw point clouds and does not need to carry out discrete quantization for 3D space, thus successfully avoiding the quantization error and realizing the structure learning in the unstructured raw unordered data. 

KPConv~\cite{thomas2019kpconv} also learned the weights of structural positions in space. However, KPConv~\cite{thomas2019kpconv} sampled the 3D point clouds for the network input and used designed linear correlation in the convolution. In comparison, ours uses the anchor-based 3D convolution (implemented by $1 \times m$ convolution kernel) to extract discrete local features, which learns geometry explicitly. In addition, our method does not require grid sampling \cite{thomas2019kpconv} and can directly deal with raw unstructured point clouds.
Mao et al.~\cite{mao2019interpolated} and Wang et al. \cite{wang2020spherical} used the interpolation based convolution method for 3D point clouds, which is consistent with the traditional discrete convolution method. However, the interpolation function is artificially designed, which can not adjust to the change of local point clouds. In contrast, our method can retain the superiority of local adaptation by spatio-temporal attentive embedding. 

Some recent works~\cite{hu2019randla,wu2020pointpwc,wang2020hierarchical} also used convolution twice, but they only used 1 $\times$ 1 convolution kernel. In this paper, different convolution kernels are introduced: 1 $\times$ 1 and $1 \times m$. The first 1 $\times$ 1 convolution organizes and encodes spatio-temporal information from unordered and irregular point clouds. Then, the structured information is extracted by the second $1 \times m$ convolution.

\subsection{Deep Learning on Dynamic 3D Point Cloud Sequences}
In MinkowskiNet~\cite{choy20194d}, 3D point cloud sequences are converted into 4D occupancy grids to deal with time sequences directly, and then the sparse 4D convolution is used. Our ASTA3DConv implements twice different convolutions, extracting more structural information than MinkowskiNet's sparse convolution. Furthermore, our method expands the searching radius with frame intervals, considering more spatio-temporally adjacent points' features.

HPLFlowNet~\cite{gu2019hplflownet} introduced DownBCL, UpBCL, and CorrBCL operations, which transferred point clouds to structured information and used the convolution, but the manual interpolation is used for data preprocessing. Unlike converting raw 3D point clouds to grids or voxels, the proposed method can directly consume dynamic 3D point cloud sequences without losing information in the data preprocessing. 

FlowNet3D~\cite{liu2019flownet3d}, based on PointNet++~\cite{qi2017pointnet++}, used flow embedding layer to associate two point clouds, and generated scene flows from the flow embedding features by the flow refinement. \cite{wu2020pointpwc} and \cite{wang2020hierarchical} used the point cost volume to associate two point clouds. The point cost volume is also based on shared MLP, implemented by 1 $\times$ 1 convolution. These methods are only for two point clouds to find the local correspondence. MeteorNet~\cite{liu2019meteornet} is one explorer of deep learning methods for dynamic 3D point cloud sequences. Direct grouping and chained-flow grouping are proposed to realize correspondence search. However, the structure of MeteorNet~\cite{liu2019meteornet} is inherited from PointNet~\cite{qi2017pointnet} and MeteorNet only uses shared MLP for interframe embedding learning. Recent PointLSTM \cite{min2020efficient} learn interframe embedding by the long short-term memory (LSTM) \cite{Donahue_2015_CVPR,hochreiter1997long}, which focuses on implicit sequence feature propagation and modeling. ASAP-Net\cite{caoasap} carries out interframe attention fusion after feature extraction of single frames, while our method directly integrates multi-frame information and then uses 3D convolution for explicit spatial feature extraction. 
P4Transformer \cite{fan2021point} uses the transformer based self-attention on embedded features to correlate the similarity of features in different frames. Our method not only uses embedded information, but also includes timestamps and spatial coordinates for attention and performs explicit 3D feature extraction through anchor-based 3D convolution.

\section{Anchor-Based Spatio-Temporal Attention 3D Convolutional Neural Network}\label{3}

A novel Anchor-based Spatio-Temporal Attention 3D Convolution (ASTA3DConv) is proposed to gather information from multi-frame point clouds structurally. Then this operation is used to build classification and segmentation networks. The anchor-based convolution is introduced in Section \ref{section:1}. The features of anchors are obtained through the spatio-temporal attentive embedding in Section \ref{section:2}. The proposed Anchor-based Spatio-Temporal Attention 3D Convolutional Neural Networks (ASTA3DCNNs) are introduced in Section \ref{section:3}.

\subsection{Anchor-Based 3D Convolution}\label{section:1}

The previous method\cite{liu2019meteornet} only uses 1 $\times$ 1 convolution, which does not make full use of the structural features of point cloud sequences. Therefore, the anchor-based convolution model is proposed in this paper. The neighborhood spatio-temporal features are first embedded to anchors, and then the proposed anchor-based convolution is applied to gather anchors' features. The anchors are expanded and connected around each convolutional central core point as shown in Fig.~\ref{fig:anc3}(b). The novel convolution gathers the neighborhood information from raw unordered and irregular point cloud sequences in a structural fashion but avoids using the indeterminate manual interpolation~\cite{mao2019interpolated,wang2020spherical}.

The overview of the proposed anchor-based spatio-temporal convolution model is shown in Fig. \ref{fig:anc3}. The input of this model includes point cloud $P_{raw}$ of sequential multiple frames and some selected points $P_{core} =\{p_i=\{x_i,f_i,t_i\}|i=1,2,...,N\}$ from $P_{raw}$. $P_{core}$ serves as central core points of the anchor-based convolution to aggregate features by Farthest Point Sampling (FPS) \cite{qi2017pointnet++} from $P_{raw}$. $x_i\in{{\mathbb{R}}^3}$ represents the 3D coordinates. $f_i\in{{\mathbb{R}}^c}$ represents the raw feature of points, and $c$ is the number of dimensions of the input feature. $t_i\in{\mathbb{N}}$ represent the timestamp (order of the frame) of the point $p_i$. The output is $P'_c =\{p'_i=\{x_i,f'_i,t_i\}|i=1,2,...,N\}$ with aggregated feature $f'_i$, and $c'$ is the number of dimensions of the output feature.

 Considering the simplest symmetrical structure of the closest-packed model of sphere \cite{hales1998overview,szpiro2003mathematics,hales2017formal,wang2020spherical},  the regular tetrahedron based 3D convolution kernel is adopted here. There are 4 vertices for a regular tetrahedron, so the built convolution kernel has only 4 learning convolution weights. $4$ virtual anchors  $A_i=\{a_i^j=\{x_i^j,t_i^j\}|j=1,2,3,4\}$ located at the 4 vertices of the regular tetrahedron can be determined based on each central core point $p_i$ as follows:
\begin{equation}\label{core_point}
	\begin{bmatrix}
		x_i^1\\
		x_i^2\\ 
		x_i^3\\
		x_i^4
	\end{bmatrix}
	=
	\begin{bmatrix}
		x_i\\ 
		x_i\\ 
		x_i\\
		x_i
	\end{bmatrix} + \Delta X\cdot S,
\end{equation} 
\begin{equation}
	t_i^j=t_i,
\end{equation}
where $x_i^j$ ($j=1,2,3,4$) and $t_i^j$ are respectively the coordinates and the timestamp of the anchor related with the core point $p_i$. $\Delta X\in \mathbb{R}$ is a scalar and represents the size of the regular tetrahedron, which decides the distance between the central point and anchors. $S$ is a 
hyperparameter set according to the defined anchor forms. For the regular tetrahedron in this paper, $S$ is defined as follows:
\begin{equation}
	S = \left[ {\begin{array}{*{20}{c}}
			{\frac{{\sqrt 2 }}{3},}&{ - \frac{{\sqrt 6 }}{3},}&{ - \frac{1}{3}} \\ 
			{\frac{{\sqrt 2 }}{3},}&{\frac{{\sqrt 6 }}{3},}&{ - \frac{1}{3}} \\ 
			{ - \frac{{2\sqrt 2 }}{3},}&{0,}&{ - \frac{1}{3}} \\ 
			{0,}&{0,}&1 
	\end{array}} \right].
\end{equation}

Each line in $S$ defines the relative position of one anchor to the central point.   
Note that anchors are virtual coordinates with virtual timestamps in the 3D space around the central core points. There may no actual points at these locations.

To this step, the coordinates and the timestamps of anchors have been obtained. With these anchors as the centers of balls, neighborhood features from different frames are gathered by ball query~\cite{qi2017pointnet++} with different radii as shown in Section \ref{section:2}. Unlike previous work~\cite{liu2019meteornet}, which directly gathers points around the core points, the proposed method gathers points based on anchors. In this way, the proposed method can make use of the structured 3D convolution on irregular 3D point cloud sequences. The successful experience of discrete convolution on images can be applied to point clouds without the need for interpolation~\cite{mao2019interpolated}.

As mentioned before, the feature extraction for central core points can be divided into two steps. The details of the first step to gather features of the neighborhood points to anchors are described in Section \ref{section:2}. The features of anchors $e_i^j$ $(j=1,2,3,4)$ are obtained by formula (\ref{Gamma}). With these features, the features of central points can be extracted through the proposed anchor-based convolution. In this convolution, the kernel size $1\times4$ corresponds to the number of anchors around each central core point in 3D space.
\begin{equation}
	f'_i=\sigma(\sum\limits_{j = 1}^4\omega_{ij}e_i^j+b_j),
\end{equation}
where $f'_i \in{{\mathbb{R}}^{c'}}$ represents the output feature of the core point $p_i$ using 3D convolution. $\omega_{ij}$ and $b_j$ are the convolution kernel parameters. $\sigma$ is the activation function, which represents the Rectified Linear Unit (ReLU) here. That is, a self-defined 3D convolution is implemented by customizing the positions of convolution kernel parameters $\omega_{ij}$ $(j=1,2,3,4)$. 

Unlike the traditional $n\times n\times n$ 3D convolution, the positions of learning parameters of our 3D convolution are defined by the anchors. As the anchors are located at the 4 vertices of the regular tetrahedrons, so the $1 \times 4$ convolution is required. Compared with the traditional 1 $\times$ 1 convolution method (Shared MLP) usually used in point clouds, the 3D convolution kernel can explicitly learn the spatial structure from 3D point clouds~\cite{mao2019interpolated,wang2020spherical}. 

\begin{figure}[t]
	\centering
	\vspace{0mm}
	\resizebox{0.60\columnwidth}{!}
	{
		\includegraphics[scale=1.0]{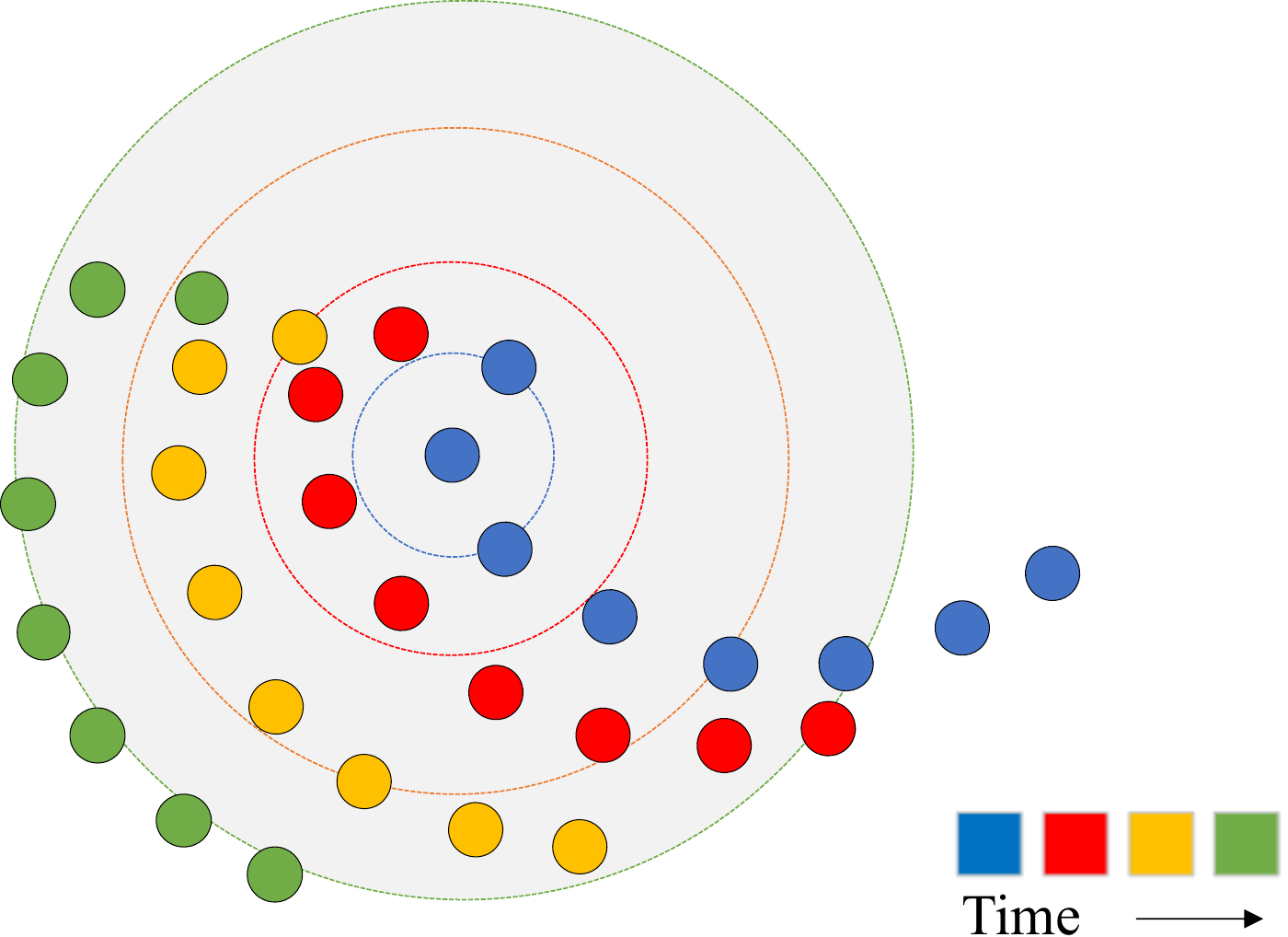}}
	\vspace{-1mm}
	\caption{Radius increases with the time frame interval. The radius for the farther frame is larger in order to receive information that matches the current frame.}
	\label{radius}
\end{figure}

\subsection{Spatio-Temporal Attentive Embedding}\label{section:2}

The anchor-based convolution in Section \ref{section:1} excavates the structural information of point cloud sequences by anchors. In this part, the method of capturing spatio-temporal information for these anchors is presented.

\begin{figure*}[t]
	\centering
	\vspace{0mm}
	\resizebox{1.0\textwidth}{!}
	{
		\includegraphics[scale=1.0]{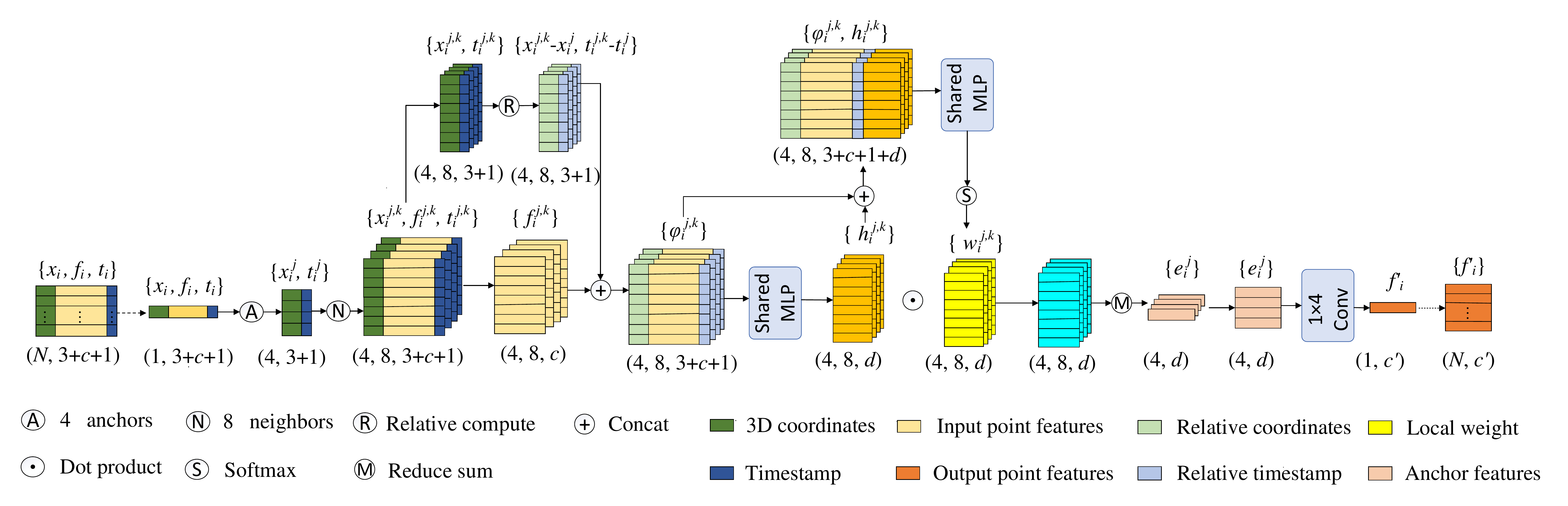}}
	\vspace{-8mm}
	\caption{The detailed computation process of our anchor-based spatio-temporal attention convolution model. 4 virutal anchors are constructed around each real point (core point) of multi-frame point clouds based on the method in Section III-A. As introduced in Section III-B, the adaptive spatial distance based query method is adopted to capture surrounding 8 points, and spatio-temporal attentive embedding is used to learn anchor features from the queried points. Then, $1 \times 4$ 3D convolution is adopted to extract structured features from the 4 anchors for each core point.} 	
	\vspace{0mm}
	\label{fig:new model}
\end{figure*}

In the proposed method, relative coordinates and timestamps of neighborhood points to anchors will be embedded in each anchor, which strengths local spatio-temporal awareness. This module also gathers raw features at the same time. The motivation for the attentive embedding is that the point clouds from different frames have different influences on the classification or segmentation. Usually, points with closer timestamps and closer distance have more significant influences. However, \cite{liu2019meteornet} only use max pooling after grouping, which ignores this important information. Therefore, the attentive embedding with spatio-temporal awareness is proposed in this paper.

To sufficiently use the information from multiple frames, selecting neighborhood points of different frames is needed. Considering the movement of objects, a natural idea is that a broader area is searched for the points as the time interval increases like~\cite{liu2019meteornet}. That is, different radii are adopted to search for the points from different frames as shown in Fig. \ref{radius}. A larger radius is adopted for far frames when gathering points so that there is a greater probability of receiving features in different frames associated with anchors.

These different radii are applied to select points around each anchor from different frames. The radii are regarded as the maximum Euclidean distances between the neighborhood points and anchors to examine whether one point can be selected for the feature extraction of anchors. The radius $R$, considering the timestamps of different frames, increases gradually with the time frame interval. A formula is used to calculate the radius:
\begin{equation}\label{r}
	R(a_i^j,p_i') = \tau \rho (\left \|  t_i^j-t_i'\right \|),
\end{equation}
where $a_i^j$ and $p_i'$ represent an anchor and an actual point around the anchor. $t_i^j$ and $t_i'$ are the timestamp of the anchor and the point separately. $\rho $ represents a monotonically increasing function with increasing time intervals, and the increasing extent is controlled by a hyperparameter $\tau$. $\left \| \cdot \right \|$ represents the absolute value of timestamp difference. Only the neighborhood points with a smaller distance from the anchors than the related radius can be selected for the feature extraction.

Through this method, $8$ neighborhood points from multiple frames $N_i^j=\{n_i^{j,k}=\{x_i^{j,k},f_i^{j,k},t_i^{j,k}\}|k=1,2,...,8\}$ around an anchor $a_i^j$ is selected for feature extraction. $x_i^{j,k}$,$f_i^{j,k}$ and $t_i^{j,k}$ represent the coordinates, raw features and timestamps of the selected points respectively.

Around the core $p_i$, there are $4$ anchors:  $A_i=\{a_i^1, a_i^2, a_i^3, a_i^4\}$. Each anchor $a_i^j$ chooses $8$ neighborhood points $\{n_i^{j,1}, n_i^{j,2}, ... ,n_i^{j,8}\}$ from multiple frames. The relative spatial coordinates and timestamps of points are used to realize spatio-temporal feature encoding:
\begin{equation}\label{equ5}
	\varphi_i^{j,k}=((x_i^{j,k} - x_i^j)\oplus \left \| t_i^{j,k} -t_i^j\right \|  \oplus f_i^{j,k}),
\end{equation}
where $x_i^j$ and $t_i^j$ represent the coordinates and timestamp of the anchor $a_i^j$. $x_i^{j,k}$, $t_i^{j,k}$ and $f_i^{j,k}$ represent the coordinates, timestamps, and features of the neighborhood points selected by this anchor. $\oplus$ represents concatenation operation. The feature $\varphi_i^{j,k}$ is used to obtain the features of anchor $a_i^j$ before the embedding based on attention:
\begin{equation}\label{equ6}
	h_i^{j,k}=MLP(\varphi_i^{j,k}),
\end{equation}
where $h_i^{j,k}\in{{\mathbb{R}}^{d}}$, and $d$ is the number of dimensions of the embedding feature. The relative spatio-temporal information, together with the features of the neighborhood points, are input to a shared MLP for multimodal information fusion and encoding. Moreover, the relative spatio-temporal information helps determine the similarity of points and influence the weights for the attentive embedding later.

\begin{figure*}[t]
	\centering
	\vspace{0mm}
	\resizebox{1.00\textwidth}{!}
	{
		\includegraphics[scale=1.0]{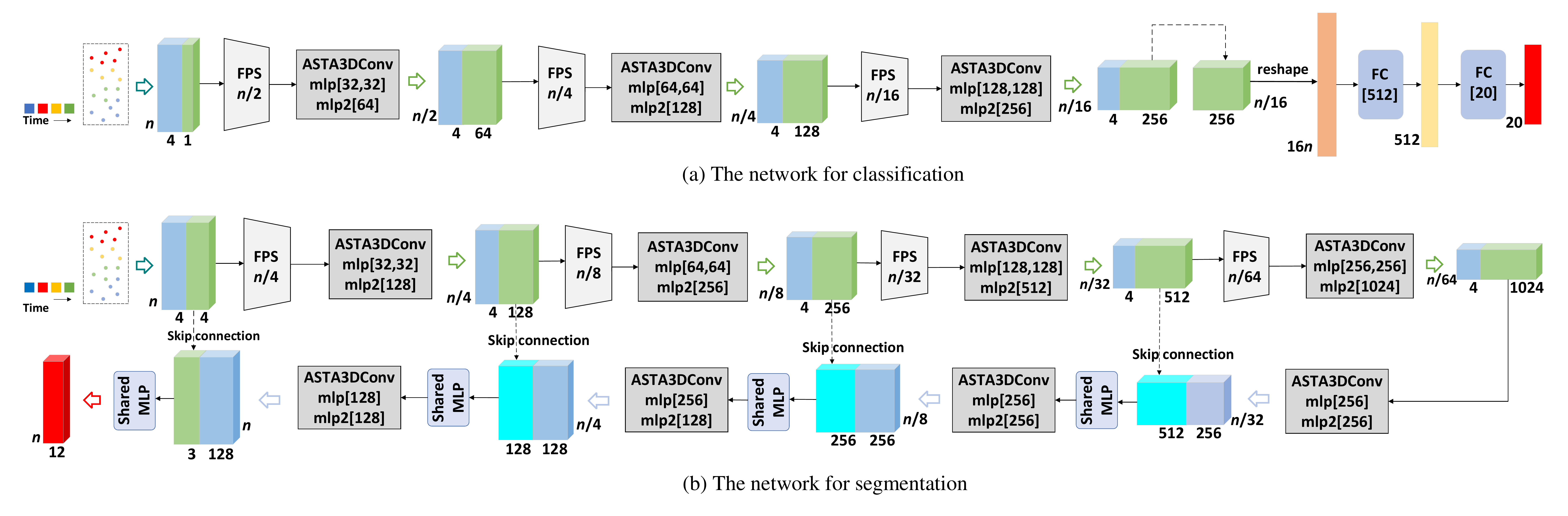}}
	\vspace{-8mm}
	\caption{The detailed diagram of our anchor-based spatio-temporal attention convolutional network. }
	\vspace{0mm}
	\label{fig:our network}
\end{figure*}

In order to aggregate the encoded information $h_i^{j,k}$ of 8 neighborhood points without losing information, we give up the max pooling used in \cite{qi2017pointnet++} and \cite{liu2019meteornet}. We expect to make use of this information in an adaptive weighted method like \cite{hu2019randla}. The gathered feature $\varphi_i^{j,k}$ introduced in equation (\ref{equ5}) helps to decide the aggregation weights of the queried points:
\begin{equation}
	w_i^{j,k}=softmax(MLP(\varphi_i^{j,k}\oplus h_i^{j,k})),
\end{equation}
where $w_i^{j,k}\in{{\mathbb{R}}^d}$ and softmax activation function is used to normalize the attention. The spatio-temporal embedding feature located at the anchor $a_i^j$ is:
\begin{equation}
	e_i^j=\sum_{k=1}^{8}{h_i^{j,k}\odot w_i^{j,k}},\label{Gamma}
\end{equation}
where $\odot$ means dot product. The proposed attentive method helps distribute attention for all the neighborhood points around anchors reasonably, which improves the accuracy of the embedding features $e_i^j$ aggregated from multiple frames. Then the features $e_i^j$ $(j=1,2,3,4)$ are aggregated to the central core point by the proposed anchor-based 3D convolution in Section \ref{section:1}.  The overview of this encoding module, together with the anchor-based model, is as shown in Fig.~\ref{fig:new model}.

By this multi-frame fusion process, the receptive field of the central core point extends to the time dimension. The most significant difference between dynamic 3D point cloud sequences and a single-frame point cloud is that the inter-frame information in the point cloud sequences dramatically contributes to the classification and segmentation results. 

As sometimes there may be no points in the local region around an anchor, a trick by setting zero is proposed to solve this problem. When an anchor can not find any points around, this anchor's feature value is set to zero. In this training step, the trainable convolution weight will have no gradient update in this local feature extraction process. Nevertheless, the trainable convolution weight can be optimized in other regions and optimized in other training steps.

\subsection{Anchor-Based Spatio-Temporal Attention 3D Convolutional Networks}\label{section:3}

In this section, we introduce two networks based on our ASTA3DConv model in detail. These two networks are designed respectively for classification and segmentation tasks from dynamic 3D point cloud sequences, shown in Fig.~\ref{fig:our network}.

\subsubsection{Network for Classification Task}This network mainly consists of 3 FPS and ASTA3DConv operations with different sizes of MLPs. The initial input of this network is 3D point clouds from multiple frames. The FPS acts as a point downsampling module for central core point selection. Then, 3D points are encoded by the ASTA3DConv. The structural features are extracted into sampled core points through the ASTA3DConv model. The final extracted features are fed into 2 Fully Connected (FC) layers, and the output is the classification results. The raw input feature only includes 1D timestamp. 
Because there are 20 classification classes for the action recognition on MSRAction3D dataset~\cite{li2010action} in Section~\ref{action}, the output of classification network is 20D vector to present the class scores of 20 classes.

\subsubsection{Network for Segmentation Task}The network of the segmentation task has an encoder-decoder architecture like U-Net~\cite{ronneberger2015u}. The encoding part of this network is similar to that of the classification network. The difference comes from the number of ASTA3DConv models and the channels of the Shared MLP. After 4 FPS layers and ASTA3DConv models, the output features are fed into the decoding part. During the decoding process, the ASTA3DConv model's output is concatenated with the encoded features from skip connections, and then the combined features are fed into a Shared MLP. After 4 ASTA3DConv models and shared MLPs, the final output provides the semantic segmentation results for all the input points. 
The input raw 4D feature includes 3D RGB color and 1D timestamp. Because there are 12 segmentation classes for the semantic segmentation on Synthia dataset~\cite{ros2016synthia} in Section~\ref{segmentation}, the output dimension of segmentation network is $n \times 12$  to present the segmentation scores of 12 classes for each point. Our classification and segmentation networks have the same input and output as MeteorNet~\cite{liu2019meteornet} because we test on the same datasets.

\section{Experiments and Evaluation}\label{4}

We designed the anchor-based spatio-temporal attention convolutional neural networks to learn the features from 3D point cloud sequences. Our networks learn from multiple frames in an adaptive weighted method and successfully extract the structural information. In this section, we will first describe the implementation details of experiments. Then our model is compared with the state-of-the-art models to show the superiority of our model. This is demonstrated in two tasks, action recognition and semantic segmentation. At last, several ablation studies are executed to analyze the contributions of our model.

\setlength{\tabcolsep}{1.2mm}
\begin{table}[t]
	\centering
	\caption{Classification accuracy on MSRAction3D dataset~\cite{li2010action}}
	\resizebox{0.98\columnwidth}{!}
	{
		\begin{tabular}{c||c|c|c}
			\toprule
			Input& Number of Frames& Method& Accuracy (\%)\\
			\hline\hline
			\noalign{\smallskip}
			Depth&Full&HON4D \cite{oreifej2013hon4d}&88.89\\
			\hline\noalign{\smallskip}
			&Full&Actionlet \cite{wang2012mining}&88.20\\
			&Full&H-HMM \cite{presti2014gesture}&89.01\\
			 &Full&Lie \cite{vemulapalli2014human}&89.48\\
			\multirow{-4}{*}{\begin{tabular}[c]{@{}c@{}}Skeleton \end{tabular}}&Full&Traj.Shape \cite{devanne20143}&92.10\\		
			\hline\noalign{\smallskip}

			& &PointNet++\cite{qi2017pointnet++}&61.61\\
			&\multirow{-2}{*}{\begin{tabular}[c]{@{}c@{}}1 \end{tabular}}&Ours w/o attention&\textbf{63.64}\\
			
			\cline{2-4}\noalign{\smallskip}
			& &MeteorNet\cite{liu2019meteornet}&78.11\\
			&&P4Transformer \cite{fan2021point}&\textbf{80.13} \\
			&\multirow{-3}{*}{\begin{tabular}[c]{@{}c@{}}4 \end{tabular}}&Ours&\textbf{80.13}\\
			\cline{2-4}\noalign{\smallskip}

			& &MeteorNet\cite{liu2019meteornet}&81.14\\
			&&P4Transformer \cite{fan2021point}&83.17 \\
			&\multirow{-3}{*}{\begin{tabular}[c]{@{}c@{}}8\end{tabular}}&Ours&\textbf{87.54}\\
			
			\cline{2-4}\noalign{\smallskip}

			& &MeteorNet\cite{liu2019meteornet}&86.53\\
			&&P4Transformer \cite{fan2021point}&87.54 \\
			&\multirow{-3}{*}{\begin{tabular}[c]{@{}c@{}}12 \end{tabular}}&Ours&\textbf{89.90}\\
			\cline{2-4}\noalign{\smallskip}
			
			&&MeteorNet\cite{liu2019meteornet}&88.21\\
			&&P4Transformer \cite{fan2021point}&89.56 \\
			&\multirow{-3}{*}{\begin{tabular}[c]{@{}c@{}}16 \end{tabular}}&Ours&\textbf{91.24}\\
			\cline{2-4}\noalign{\smallskip}
			& &MeteorNet\cite{liu2019meteornet}&88.50\\ &
			&Body Surface Context \cite{song2014body}&90.36\\
			&&P4Transformer \cite{fan2021point}&90.94 \\
			&&PointLSTM \cite{min2020efficient}&92.29\\
			\multirow{-22}{*}{\begin{tabular}[c]{@{}c@{}}Points \end{tabular}}&\multirow{-5}{*}{\begin{tabular}[c]{@{}c@{}}24 (Full) \end{tabular}}&Ours&\textbf{93.03}\\
			\bottomrule
			
	\end{tabular}}
	\label{table:1}
\end{table}

\setlength{\tabcolsep}{1.3mm}{
	\begin{table*}[t]
		\centering
		\vspace{2mm}
		\caption{Semantic segmentation results on the Synthia dataset~\cite{ros2016synthia}}
		\begin{tabular}{c|c|cccccccccccc|c}
			\toprule
			Number of Frames&Method&Blding&Road&Sdwlk&Fence&Vegitn&Pole&Car&T.sign&Pdstr&Bicyc&Lane&T.light&mIoU\\
			\hline\hline
			\noalign{\smallskip}
			&4D MinkNet14\cite{choy20194d}&89.39&97.68&69.43&86.52&98.11&97.26&93.50&\textbf{79.45}&\textbf{92.27}&0.00&44.61&66.69&76.24\\
			&PointNet++\cite{qi2017pointnet++}&96.88&97.72&86.20&92.75&97.12&97.09&90.85&66.87&78.64&0.00&72.93&75.17&79.35\\
			\multirow{-3}{*}{\begin{tabular}[c]{@{}c@{}}1 \end{tabular}}&Ours w/o attention&\textbf{98.35}&\textbf{98.72}&\textbf{93.28}&\textbf{96.56}&\textbf{98.84}&\textbf{97.91}&\textbf{95.35}&75.36&82.81&0.00&\textbf{77.05}&\textbf{80.78}&\textbf{82.92}\\
			\hline\hline
			\noalign{\smallskip}

			&MeteorNet\cite{liu2019meteornet}&97.65&97.83&90.03&94.06&97.41&97.79&94.15&\textbf{82.01}&79.14&0.00&72.59&77.92&81.72\\
			\multirow{-2}{*}{\begin{tabular}[c]{@{}c@{}}2 \end{tabular}}&Ours&\textbf{98.52}&\textbf{98.80}&\textbf{94.46}&\textbf{96.80}&\textbf{99.06}&\textbf{98.46}&\textbf{96.31}&81.04&\textbf{88.23}&0.00&\textbf{77.68}&\textbf{81.72}&\textbf{84.26}\\
			
			\hline\hline
			\noalign{\smallskip}
			&4D MinkNet14\cite{choy20194d}&90.13&98.26&73.47&87.19&\textbf{99.10}&97.50&94.01&79.04&\textbf{92.62}&0.00&50.01&68.14&77.46\\
			&MeteorNet\cite{liu2019meteornet}&98.10&97.72&88.65&94.00&97.98&97.65&93.83&84.07&80.90&0.00&71.14&77.60&81.80\\
			&ASAP-Net\cite{caoasap}&97.67 &98.15 &89.85 &95.50& 97.12& 97.59 &94.90& 80.97 &86.08 &0.00 &74.66& 77.51&82.73\\
			&P4Transformer\cite{fan2021point}& 
			96.73 &98.35 &94.03 &95.23&98.28& 98.01 &\textbf{95.60}& 81.54 &85.18 &0.00 &75.95& 79.07&83.16\\

			\multirow{-5}{*}{\begin{tabular}[c]{@{}c@{}}3 \end{tabular}}&Ours& \textbf{98.23} &\textbf{98.78}  & \textbf{95.38}& \textbf{96.38} & 98.61&\textbf{98.56} &95.39& \textbf{84.14}& 87.78&0.00 & \textbf{78.46}& \textbf{85.49} &\textbf{84.77} \\
			\bottomrule

		\end{tabular}

		\label{table:2}
	\end{table*}
}

\subsection{Implementation}
For action recognition, the initial learning rate is set as 0.001. For semantic segmentation, the initial learning rate is set as 0.0016. For both of them, the learning rate decreases by 0.7 every $200{,}000$ steps. The number of input points is the sum of the points from multiple frames. For action classification and semantic segmentation, the number of points in a single frame is  $2{,}048$ and $8{,}192$, respectively. The radii for querying neighborhood points for anchors are related to the timestamps of the points and the density of point clouds in the current feature extraction level. After each FPS, we will double the radius to get a fixed number of neighborhood points in the point clouds before this FPS. As [0.5,0.6] and [0.98,1.0] are used as the initial radius distribution in \cite{liu2019meteornet}, we use these as the basic parameters. The initial radii for different frames are evenly distributed between $a\times0.5$ and $a\times0.6$ according to the timestamps for action classification. For semantic segmentation, the radii are between $a\times0.98$ and $a\times1.0$. $a$ is an adjustment coefficient determined by experiments. $a$ is $0.25$ for action recognition and $1.1$ for semantic segmentation. Furthermore, the smallest radius is set to be equal to the distance between an anchor and its related central core point. BatchNorm~\cite{ioffe2015batch} is used following each ASTA3DConv operation, and Adam optimizer~\cite{kingma2014adam} is adopted. Our method consumes the dynamic 3D point cloud sequences. Therefore, MSRAction3D dataset~\cite{li2010action} and Synthia dataset~\cite{ros2016synthia} are preprocessed to generate dynamic 3D point cloud sequences as input for training and testing. The experiments for action recognition are all performed on a single RTX 2080Ti GPU. The experiments for semantic segmentation are all on a single Titan RTX GPU.

\subsection{Action Recognition}\label{action}

\begin{figure}[t]
	\centering
	\vspace{0mm}

	\resizebox{1.00\columnwidth}{!}
	{
		\subfigure{\includegraphics[width=0.24\textwidth,height=0.38\textwidth]{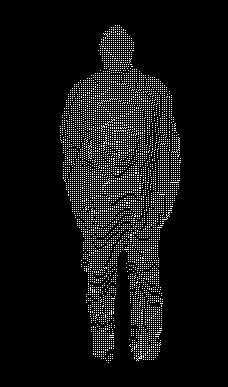}}
		\subfigure{\includegraphics[width=0.24\textwidth,height=0.38\textwidth]{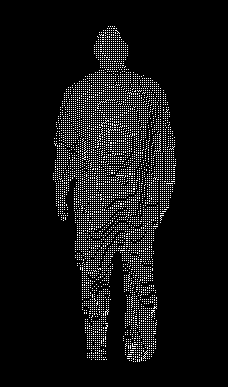}}
		\subfigure{\includegraphics[width=0.24\textwidth,height=0.38\textwidth]{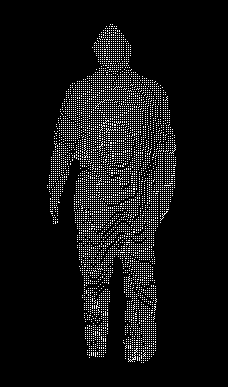}}
		\subfigure{\includegraphics[width=0.24\textwidth,height=0.38\textwidth]{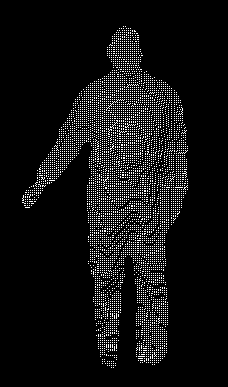}}
		\subfigure{\includegraphics[width=0.24\textwidth,height=0.38\textwidth]{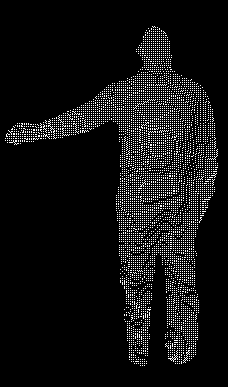}}
		\subfigure{\includegraphics[width=0.24\textwidth,height=0.38\textwidth]{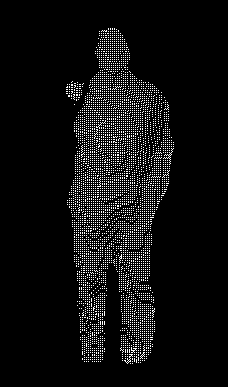}}
		\subfigure{\includegraphics[width=0.24\textwidth,height=0.38\textwidth]{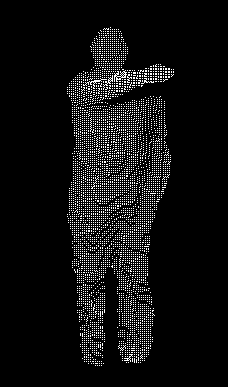}}
		\subfigure{\includegraphics[width=0.24\textwidth,height=0.38\textwidth]{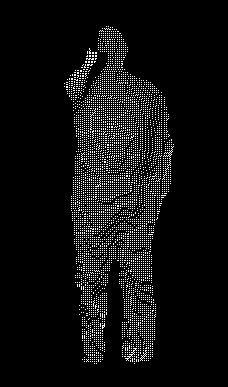}}
		\subfigure{\includegraphics[width=0.24\textwidth,height=0.38\textwidth]{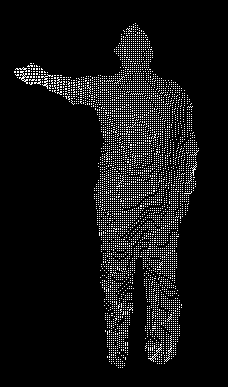}}
		\subfigure{\includegraphics[width=0.24\textwidth,height=0.38\textwidth]{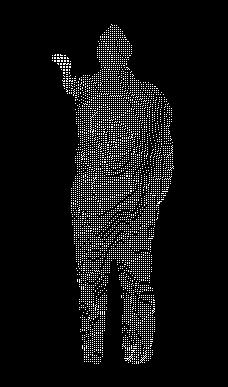}}
		\subfigure{\includegraphics[width=0.24\textwidth,height=0.38\textwidth]{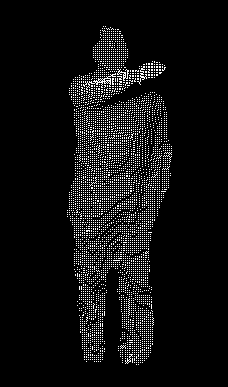}}
		\subfigure{\includegraphics[width=0.24\textwidth,height=0.38\textwidth]{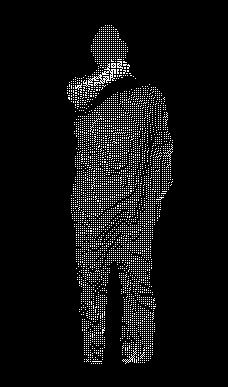}}
		\subfigure{\includegraphics[width=0.24\textwidth,height=0.38\textwidth]{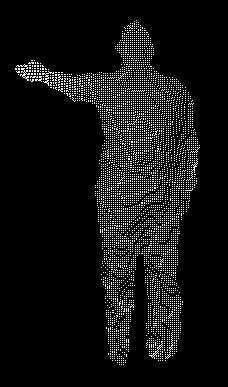}}
		\subfigure{\includegraphics[width=0.24\textwidth,height=0.38\textwidth]{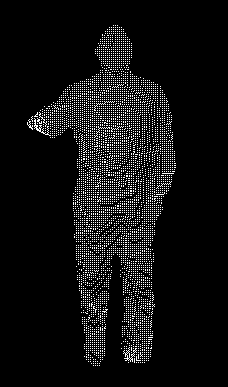}}
		\subfigure{\includegraphics[width=0.24\textwidth,height=0.38\textwidth]{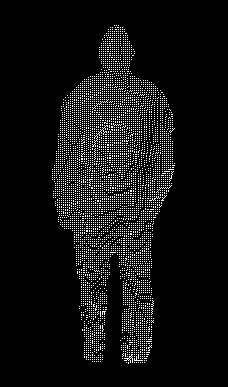}}
	}\\\vspace{0.1cm}
	
	\resizebox{1.00\columnwidth}{!}
	{
		\subfigure{\includegraphics[width=0.24\textwidth,height=0.38\textwidth]{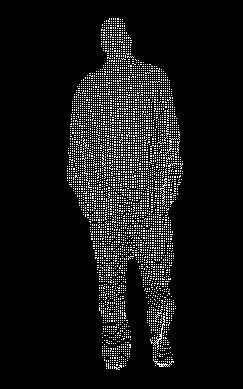}}
		\subfigure{\includegraphics[width=0.24\textwidth,height=0.38\textwidth]{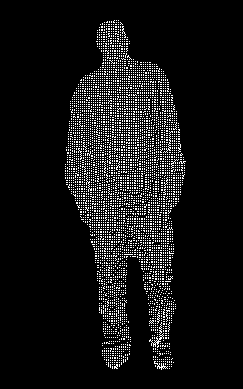}}
		\subfigure{\includegraphics[width=0.24\textwidth,height=0.38\textwidth]{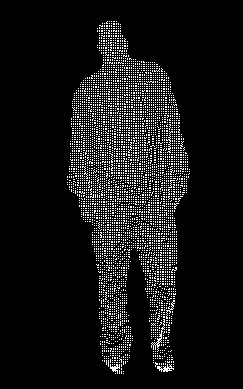}}
		\subfigure{\includegraphics[width=0.24\textwidth,height=0.38\textwidth]{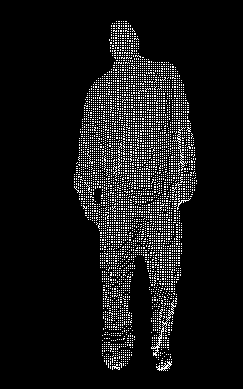}}
		\subfigure{\includegraphics[width=0.24\textwidth,height=0.38\textwidth]{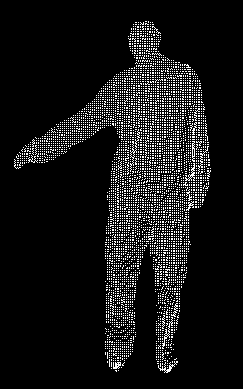}}
		\subfigure{\includegraphics[width=0.24\textwidth,height=0.38\textwidth]{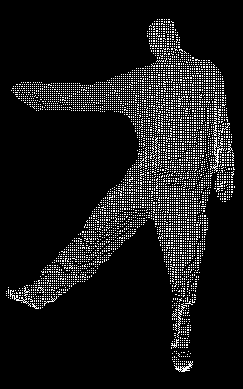}}
		\subfigure{\includegraphics[width=0.24\textwidth,height=0.38\textwidth]{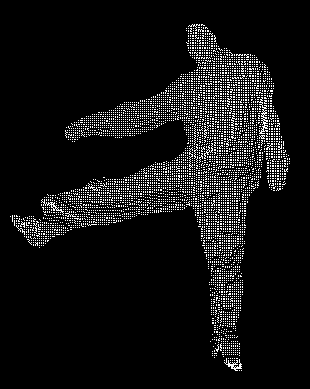}}
		\subfigure{\includegraphics[width=0.24\textwidth,height=0.38\textwidth]{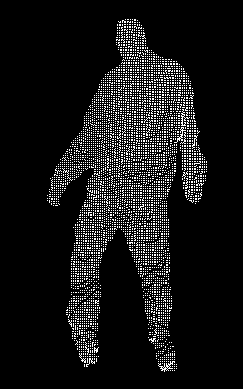}}
		\subfigure{\includegraphics[width=0.24\textwidth,height=0.38\textwidth]{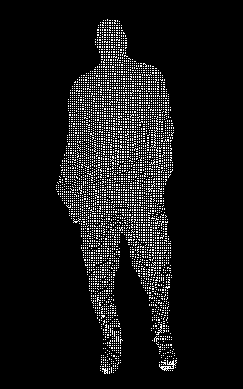}}
		\subfigure{\includegraphics[width=0.24\textwidth,height=0.38\textwidth]{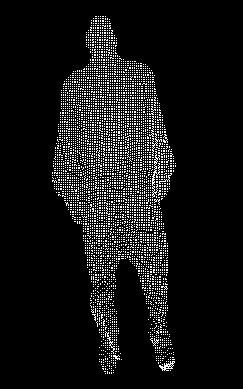}}
		\subfigure{\includegraphics[width=0.24\textwidth,height=0.38\textwidth]{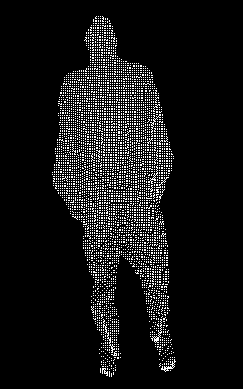}}
		\subfigure{\includegraphics[width=0.24\textwidth,height=0.38\textwidth]{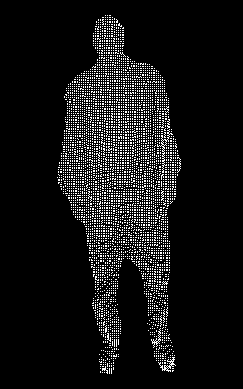}}
		\subfigure{\includegraphics[width=0.24\textwidth,height=0.38\textwidth]{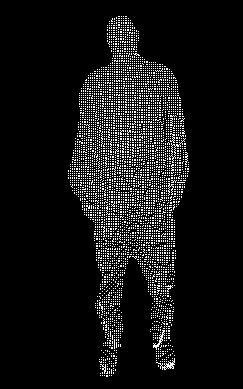}}
		\subfigure{\includegraphics[width=0.24\textwidth,height=0.38\textwidth]{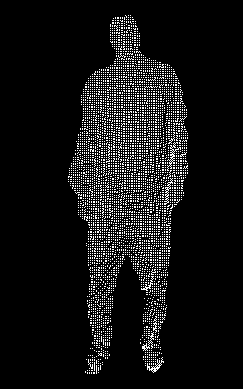}}
		\subfigure{\includegraphics[width=0.24\textwidth,height=0.38\textwidth]{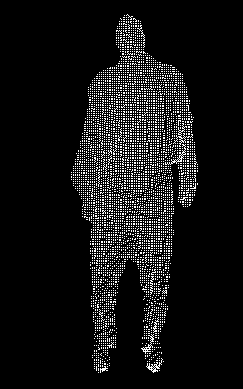}}
	}\\\vspace{0.1cm}

	\resizebox{1.00\columnwidth}{!}
	{
		\subfigure{\includegraphics[width=0.24\textwidth,height=0.38\textwidth]{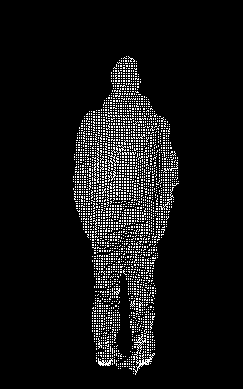}}
		\subfigure{\includegraphics[width=0.24\textwidth,height=0.38\textwidth]{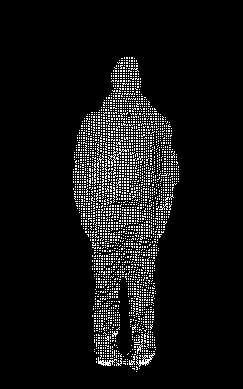}}
		\subfigure{\includegraphics[width=0.24\textwidth,height=0.38\textwidth]{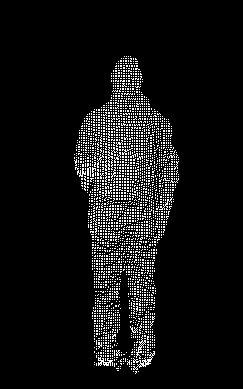}}
		\subfigure{\includegraphics[width=0.24\textwidth,height=0.38\textwidth]{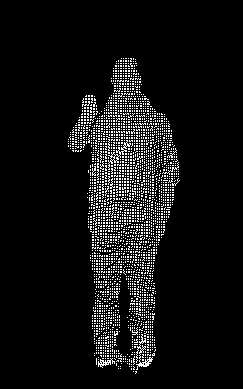}}
		\subfigure{\includegraphics[width=0.24\textwidth,height=0.38\textwidth]{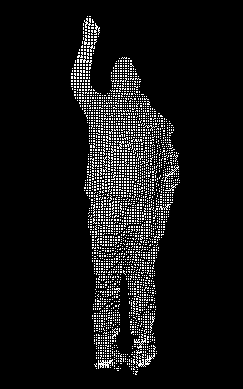}}
		\subfigure{\includegraphics[width=0.24\textwidth,height=0.38\textwidth]{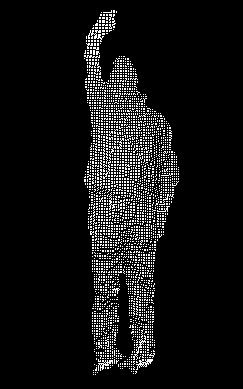}}
		\subfigure{\includegraphics[width=0.24\textwidth,height=0.38\textwidth]{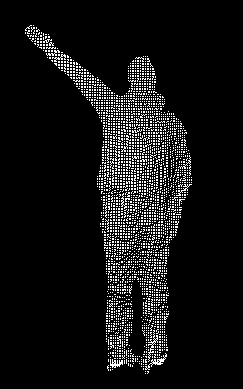}}
		\subfigure{\includegraphics[width=0.24\textwidth,height=0.38\textwidth]{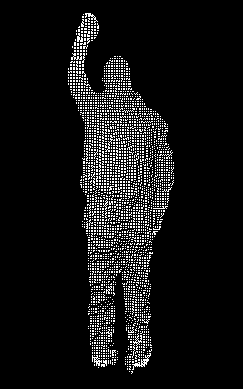}}
		\subfigure{\includegraphics[width=0.24\textwidth,height=0.38\textwidth]{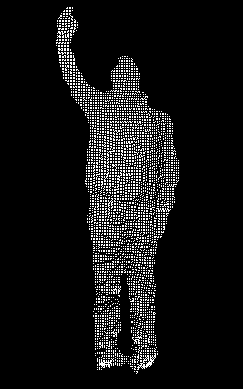}}
		\subfigure{\includegraphics[width=0.24\textwidth,height=0.38\textwidth]{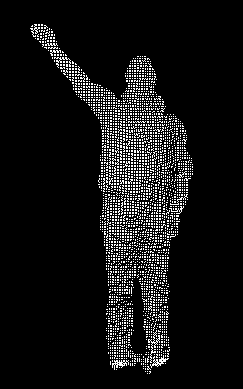}}
		\subfigure{\includegraphics[width=0.24\textwidth,height=0.38\textwidth]{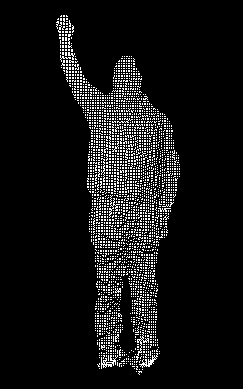}}
		\subfigure{\includegraphics[width=0.24\textwidth,height=0.38\textwidth]{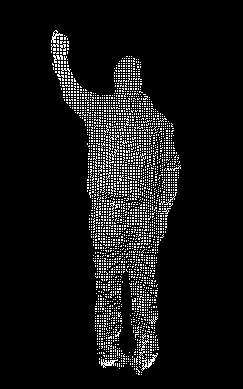}}
		\subfigure{\includegraphics[width=0.24\textwidth,height=0.38\textwidth]{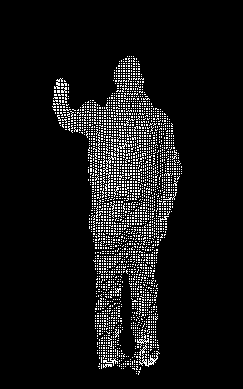}}
		\subfigure{\includegraphics[width=0.24\textwidth,height=0.38\textwidth]{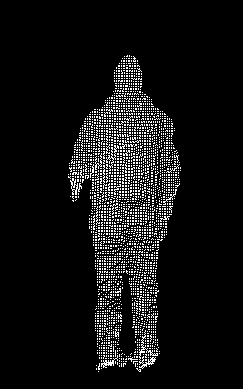}}
		\subfigure{\includegraphics[width=0.24\textwidth,height=0.38\textwidth]{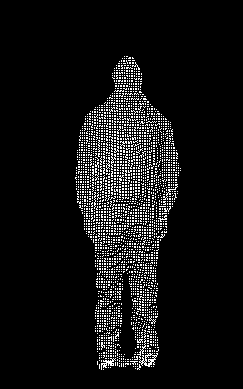}}
		
	}\\\vspace{0.1cm}
	
	\resizebox{1.00\columnwidth}{!}
	{
		\subfigure{\includegraphics[width=0.24\textwidth,height=0.38\textwidth]{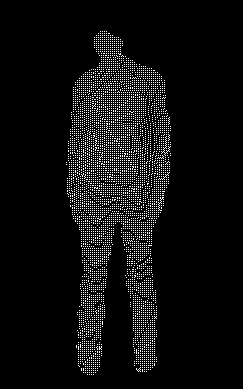}}
		\subfigure{\includegraphics[width=0.24\textwidth,height=0.38\textwidth]{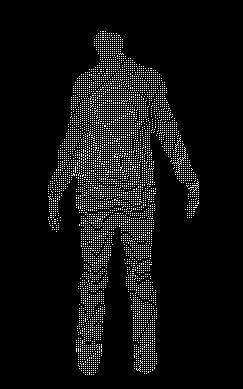}}
		\subfigure{\includegraphics[width=0.24\textwidth,height=0.38\textwidth]{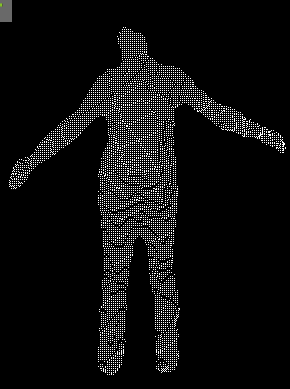}}
		\subfigure{\includegraphics[width=0.24\textwidth,height=0.38\textwidth]{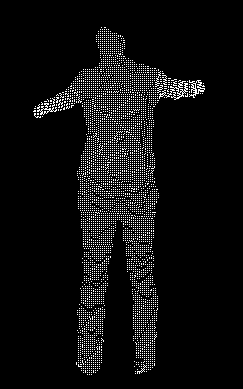}}
		\subfigure{\includegraphics[width=0.24\textwidth,height=0.38\textwidth]{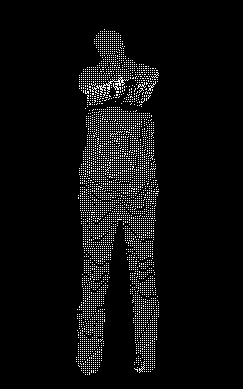}}
		\subfigure{\includegraphics[width=0.24\textwidth,height=0.38\textwidth]{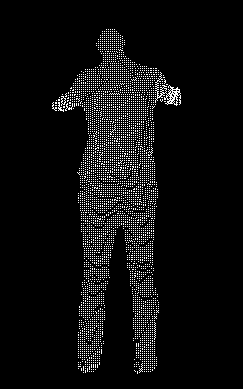}}
		\subfigure{\includegraphics[width=0.24\textwidth,height=0.38\textwidth]{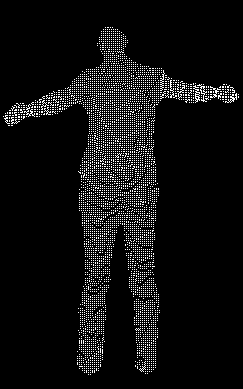}}
		\subfigure{\includegraphics[width=0.24\textwidth,height=0.38\textwidth]{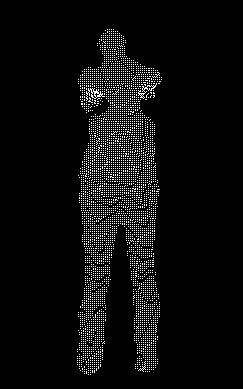}}
		\subfigure{\includegraphics[width=0.24\textwidth,height=0.38\textwidth]{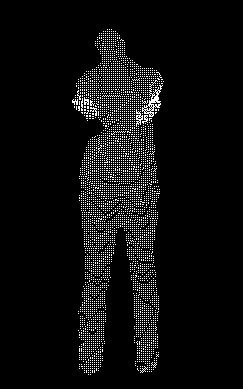}}
		\subfigure{\includegraphics[width=0.24\textwidth,height=0.38\textwidth]{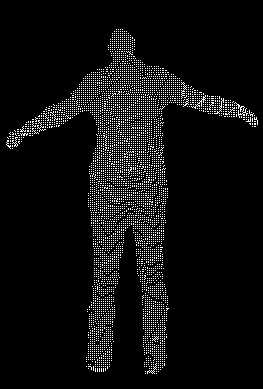}}
		\subfigure{\includegraphics[width=0.24\textwidth,height=0.38\textwidth]{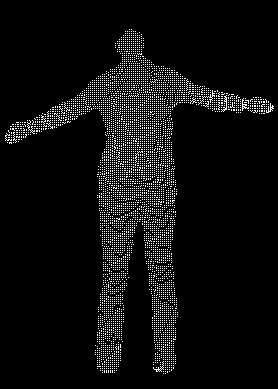}}
		\subfigure{\includegraphics[width=0.24\textwidth,height=0.38\textwidth]{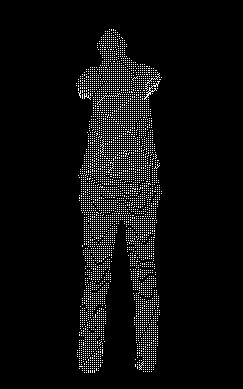}}
		\subfigure{\includegraphics[width=0.24\textwidth,height=0.38\textwidth]{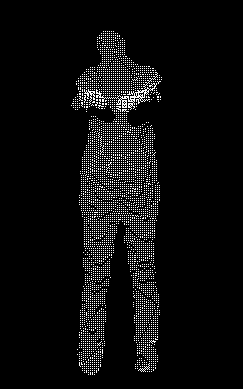}}
		\subfigure{\includegraphics[width=0.24\textwidth,height=0.38\textwidth]{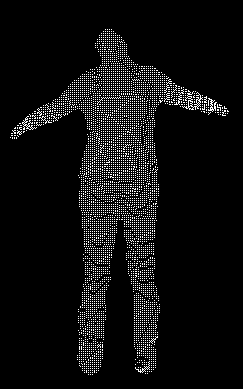}}
		\subfigure{\includegraphics[width=0.24\textwidth,height=0.38\textwidth]{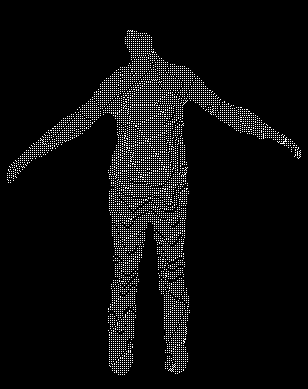}}
		
	}\\\vspace{0.1cm}

	\resizebox{1.00\columnwidth}{!}
	{
		\subfigure{\includegraphics[width=0.24\textwidth,height=0.38\textwidth]{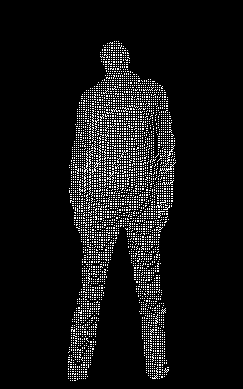}}
		\subfigure{\includegraphics[width=0.24\textwidth,height=0.38\textwidth]{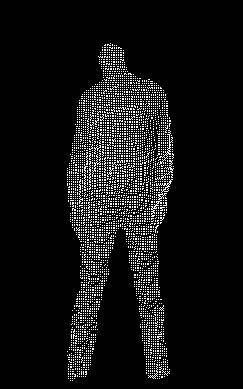}}
		\subfigure{\includegraphics[width=0.24\textwidth,height=0.38\textwidth]{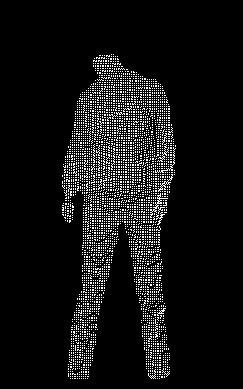}}
		\subfigure{\includegraphics[width=0.24\textwidth,height=0.38\textwidth]{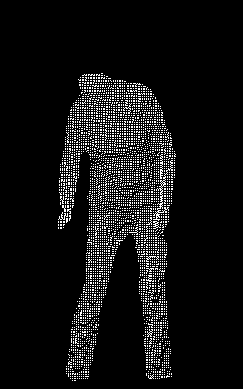}}
		\subfigure{\includegraphics[width=0.24\textwidth,height=0.38\textwidth]{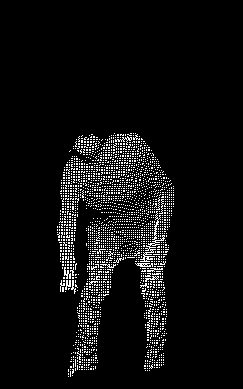}}
		\subfigure{\includegraphics[width=0.24\textwidth,height=0.38\textwidth]{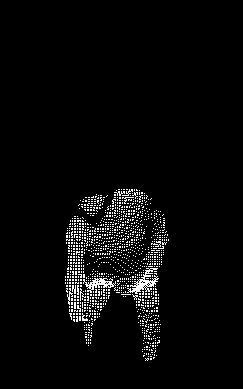}}
		\subfigure{\includegraphics[width=0.24\textwidth,height=0.38\textwidth]{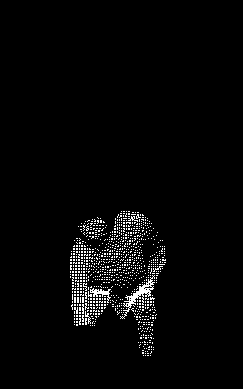}}
		\subfigure{\includegraphics[width=0.24\textwidth,height=0.38\textwidth]{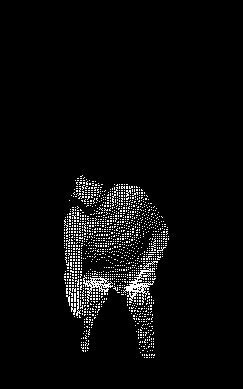}}
		\subfigure{\includegraphics[width=0.24\textwidth,height=0.38\textwidth]{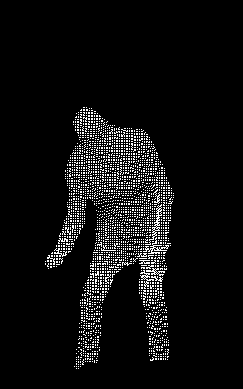}}
		\subfigure{\includegraphics[width=0.24\textwidth,height=0.38\textwidth]{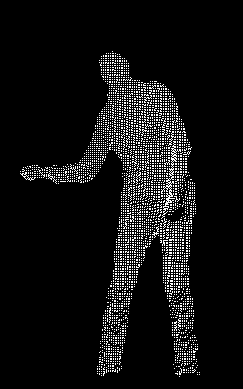}}
		\subfigure{\includegraphics[width=0.24\textwidth,height=0.38\textwidth]{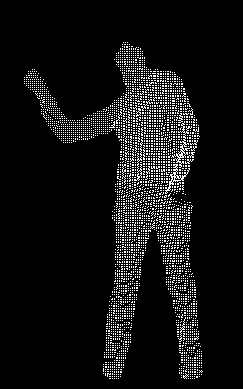}}
		\subfigure{\includegraphics[width=0.24\textwidth,height=0.38\textwidth]{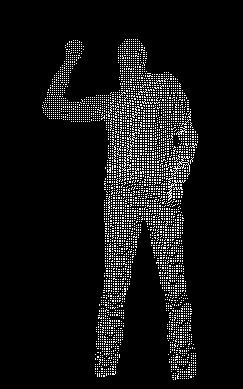}}
		\subfigure{\includegraphics[width=0.24\textwidth,height=0.38\textwidth]{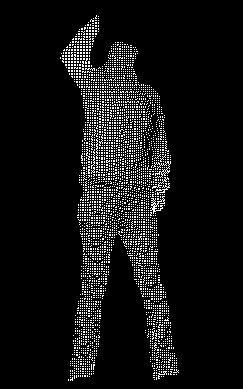}}
		\subfigure{\includegraphics[width=0.24\textwidth,height=0.38\textwidth]{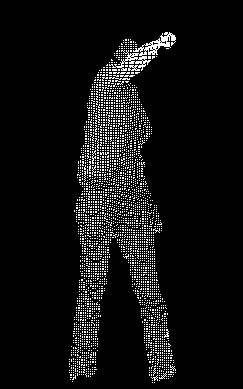}}
		\subfigure{\includegraphics[width=0.24\textwidth,height=0.38\textwidth]{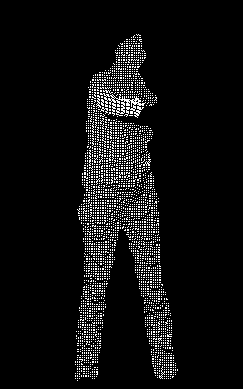}}		
	}\\\vspace{0.0cm}
	\caption{Some samples of MSRAction3D dataset~\cite{li2010action}. From the top to the bottom: horizontal arm wave, side kick, high arm wave, hand clap, pickup \& throw.}\label{fig:action}
\end{figure}

The proposed classification network is applied to action recognition on MSRAction3D dataset~\cite{li2010action}. This dataset contains 567 Kinect depth map sequences from 10 people with 20 actions.
Dynamic 3D point cloud sequences are reconstructed by these depth map sequences. Some action examples are shown in Fig.~\ref{fig:action}. The training set and test set division is the same as the previous works~\cite{wang2012mining,liu2019meteornet}.

The classification results are shown in Table~\ref{table:1}. The evaluation metric is the average classification accuracy on the test set. We compare our method with the depth based methods~\cite{oreifej2013hon4d}, skeleton based method~\cite{wang2012mining,presti2014gesture,vemulapalli2014human,devanne20143}, and the point cloud based methods~\cite{song2014body,qi2017pointnet++,liu2019meteornet,min2020efficient,fan2021point}. Note that PointNet++~\cite{qi2017pointnet++} is a single point cloud learning method, while MetoerNet~\cite{liu2019meteornet}, PointLSTM \cite{min2020efficient}, and P4Transformer \cite{fan2021point} are the learning based methods that consumes point cloud sequences, like ours. Therefore, the comparison between ours and \cite{liu2019meteornet,min2020efficient,fan2021point} is mostly fair. As shown in Table~\ref{table:1}, the result of ours with 12-frame input has even exceeded the 24-frame results of MetoerNet~\cite{liu2019meteornet}. Finally, with 24-frame input, our method realizes more than 4\% improvement over MetoerNet~\cite{liu2019meteornet}. In addition, our method surpasses the traditional method \cite{song2014body} consuming point cloud sequences, which is also a spatio-temporal scheme. Our method even surpasses the recent LSTM and transformer based methods \cite{min2020efficient,fan2021point}, which shows that our spatio-temporal awareness attention learning even exceeds the mainstream sequence model, LSTM, and the mainstream attention model, transformer, in 3D point cloud sequence tasks and achieves state of the art.

The customized 3D structured anchor distribution explicitly extracts the structural features from the point cloud sequences. Thereby, a more structural and comprehensive understanding of the sequential action improved the experimental results.

\begin{table}[t]
	\centering
	\caption{Ablation studies of action recognition on MSRAction3D dataset~\cite{li2010action}. Classification accuracy (\%) is used for the evaluation.}
	\resizebox{0.90\columnwidth}{!}
	{
		\begin{tabular}{c||c|c|c}
			\toprule
			\ Number of Frames&MeteorNet~\cite{liu2019meteornet}&Ours w/o attention&Ours (full) \\
			\hline\hline
			\noalign{\smallskip}
			4&78.11&78.45&\textbf{80.13}\\
			8&81.14&86.20&\textbf{87.54}\\
			12&86.53&89.23&\textbf{89.90}\\
			16&88.21&90.91&\textbf{91.25}\\
			24&88.50&91.29&\textbf{93.03}\\
			\bottomrule
	\end{tabular}}
	\label{table:3}
\end{table}

\begin{table*}[t]
	\centering
	\caption{Ablation studies of semantic segmentation on the Synthia dataset~\cite{ros2016synthia}}
	\begin{tabular}{c|c|cccccccccccc|c}
		\toprule
		Number of Frames&Method&Blding&Road&Sdwlk&Fence&Vegitn&Pole&Car&T.sign&Pdstr&Bicyc&Lane&T.light&mIoU\\
		\hline\hline
		\noalign{\smallskip}
		
		&MeteorNet\cite{liu2019meteornet}&97.65&97.83&90.03&94.06&97.41&97.79&94.15&\textbf{82.01}&79.14&0.00&72.59&77.92&81.72\\

		&Ours w/o attention&98.37&98.79&94.26&96.56&98.98&98.12&95.88&78.96&87.66&0.00&\textbf{77.70}&\textbf{81.86}&83.93\\
		
		\multirow{-3}{*}{\begin{tabular}[c]{@{}c@{}}2 \end{tabular}}&Ours (full, with attention)&\textbf{98.52}&\textbf{98.80}&\textbf{94.46}&\textbf{96.80}&\textbf{99.06}&\textbf{98.46}&\textbf{96.31}&81.04&\textbf{88.23}&0.00&77.68&81.72&\textbf{84.26}\\
		\hline\hline
		\noalign{\smallskip}
		
		&MeteorNet\cite{liu2019meteornet}&98.10&97.72&88.65&94.00&97.98&97.65&93.83&84.07&80.90&0.00&71.14&77.60&81.80\\
		
		&Ours w/o attention&\textbf{98.39}&\textbf{98.80}&94.50&\textbf{96.96}&\textbf{99.08}&98.45&\textbf{96.04}&\textbf{84.82}&85.17&0.00&78.03&81.54&84.31\\
		
		\multirow{-3}{*}{\begin{tabular}[c]{@{}c@{}}3 \end{tabular}}&Ours (full, with attention)& 98.23 &98.78  & \textbf{95.38}& 96.38 & 98.61&\textbf{98.56} &95.39& 84.14& \textbf{87.78}&0.00 & \textbf{78.46}& \textbf{85.49} &\textbf{84.77} \\
		\bottomrule
	\end{tabular}
	\label{table:4}
\end{table*}

\subsection{Semantic Segmentation}\label{segmentation}

Our network for semantic segmentation is tested on the Synthia dataset~\cite{ros2016synthia}. Synthia dataset~\cite{ros2016synthia} is about driving scenarios and is used for semantic segmentation and related scene understanding tasks. The original video sequences are stereo RGBD images generated by 4 cameras located on the top of a moving car. In our task, RGB images and depth maps are utilized to generate the point cloud sequences. Overall, 6 video sequences in 9 different weather environments are preprocessed to create dynamic 3D point cloud sequences. For each frame of dynamic point cloud sequences, a cube with a limit of $50m \times 50m \times 50m$ is built, where the moving car is in the center. Inside each frame, the FPS is applied to obtain $8{,}192$ points. The train/validation/test split is set as follows: Sequences 1-4 except for spring, sunset, and fog conditions are set as the training set. Sequence 5 is used as the validation set. Sunset and spring scenes in sequence 6 are used as the test set. Among the input dynamic 3D point cloud sequences, there are $19{,}888$ frames for the training set, 815 frames for the validation set, and 1,886 frames for the test set.

\begin{figure}[t]
	\centering
	\vspace{0mm}
	
	\resizebox{1.00\columnwidth}{!}
	{
		\setlength{\tabcolsep}{8mm}{
			\begin{tabular}{ccc}
				{\small RGB input}  &    {\small Ground truth}  &   {\small Our prediction}
	\end{tabular}}}
	
	\resizebox{1.00\columnwidth}{!}
	{
		\subfigure{\includegraphics[width=0.3\textwidth,height=0.2\textwidth]{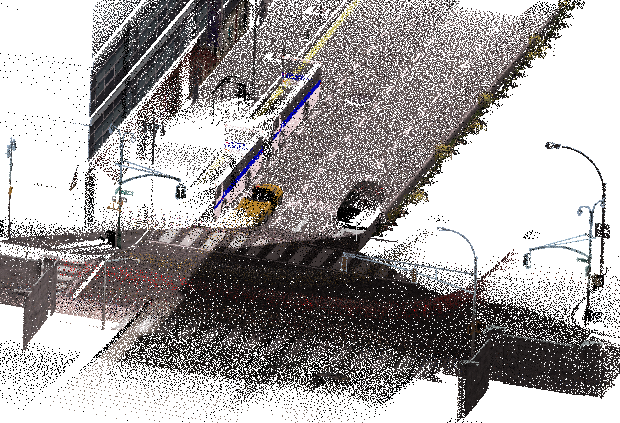}}
		\subfigure{\includegraphics[width=0.3\textwidth,height=0.2\textwidth]{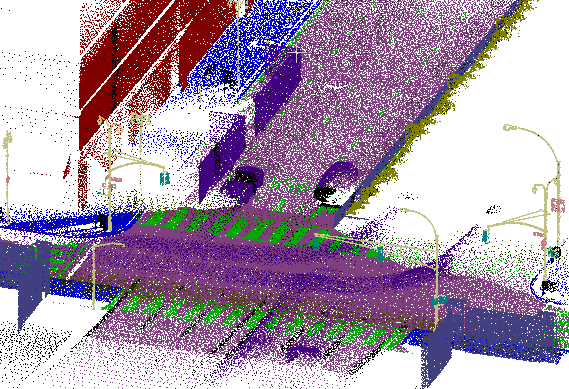}}
		\subfigure{\includegraphics[width=0.3\textwidth,height=0.2\textwidth]{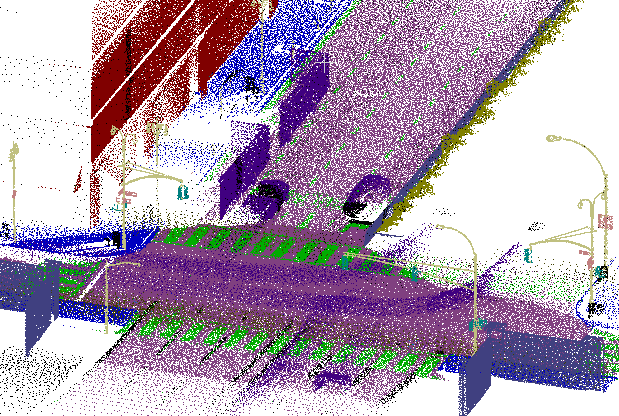}}}\\\vspace{0.1cm}

	\resizebox{1.00\columnwidth}{!}
	{
		\subfigure{\includegraphics[width=0.3\textwidth,height=0.2\textwidth]{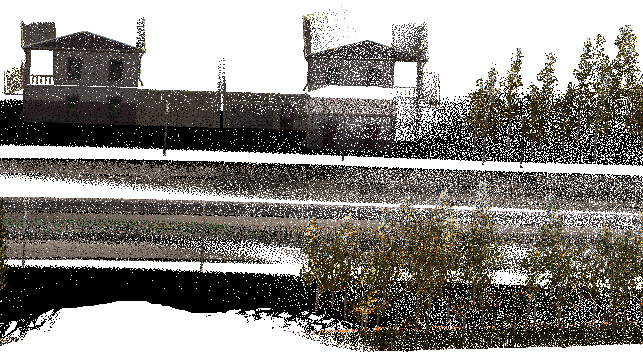}}
		\subfigure{\includegraphics[width=0.3\textwidth,height=0.2\textwidth]{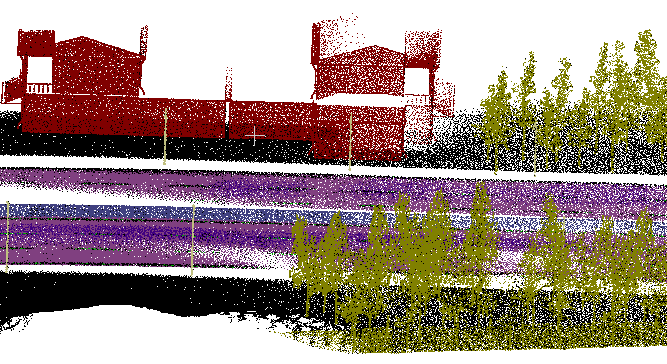}}
		\subfigure{\includegraphics[width=0.3\textwidth,height=0.2\textwidth]{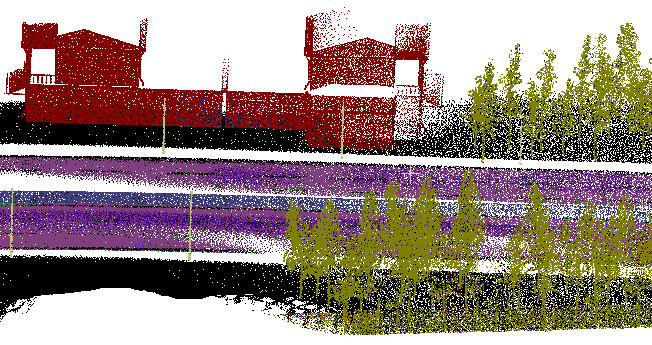}}}\\\vspace{0.1cm}
	
	\resizebox{1.00\columnwidth}{!}
	{
		\subfigure{\includegraphics[width=0.3\textwidth,height=0.2\textwidth]{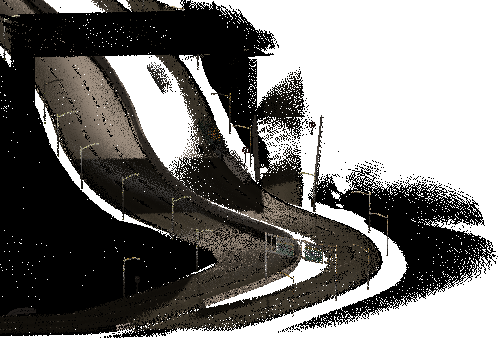}}
		\subfigure{\includegraphics[width=0.3\textwidth,height=0.2\textwidth]{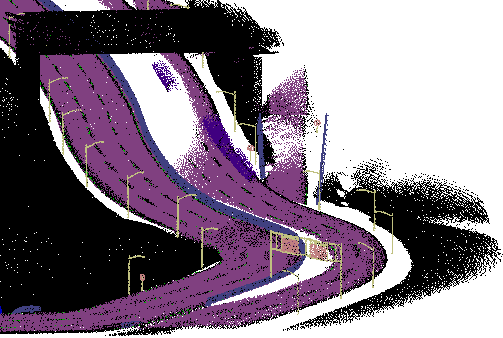}}
		\subfigure{\includegraphics[width=0.3\textwidth,height=0.2\textwidth]{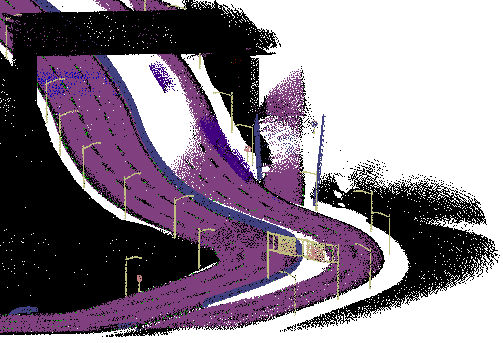}}}\\\vspace{-2mm}
	\caption{Visualized results on Synthia dataset~\cite{ros2016synthia}. Our method has a good prediction in many details, such as street lamps, roads, and houses.}\label{fig:visu}
	
\end{figure}

The results are listed in Table \ref{table:2}. The evaluation metrics are based on per-class and mean Intersection-over-Union (IoU). Our method is compared with 4D MinkNet14~\cite{choy20194d}, PointNet++\cite{qi2017pointnet++} and recent work MeteorNet~\cite{liu2019meteornet}. The model is tested by changing the number of input frames. Among the baselines, MeteorNet\cite{liu2019meteornet}, ASAP-Net\cite{caoasap}, and P4Transformer\cite{fan2021point} are the methods that perform semantic segmentation on raw 3D point cloud sequences, which is most similar to our method. Our method has the best results with different numbers of input frames and achieves state-of-the-art performance.
Some visualized results are presented in Fig.~\ref{fig:visu}. It can be found that the proposed method achieves accurate semantic prediction for most points.

\subsection{Ablation Study}
The ablation studies are executed to demonstrate the proposed contributions in this paper. The experiment settings are same with Section~\ref{action} and Section~\ref{segmentation} except for ablation components.

\subsubsection{Anchor-Based 3D Convolution}
One of the essential components of our model is the anchor-based 3D convolution. Through the anchors, central core points are able to excavate features in two steps with a structural method. To show the effect of anchors, we compare the result of ours without the attentive embedding and that of MeteorNet~\cite{liu2019meteornet} on action recognition task in Table \ref{table:3}. We find that the anchors help improve the accuracy at least 2.5\% for all the number of input frames except 4 frames. For 4 frames as input, the accuracy of our model with anchors (w/o attention) also exceeds MeteorNet~\cite{liu2019meteornet}. There is an explanation for this. Most actions require a more extended period to be correctly classified. For the input with fewer frames, it is hard to get a significant improvement in accuracy even though the anchor-based method is applied.
For semantic segmentation, structure information also exists in a single frame. Therefore, as shown in Table~\ref{table:4}, the performance gain is enough significant for fewer frame inputs using the anchor-based method.

MeteorNet~\cite{liu2019meteornet} will degenerate into PointNet++~\cite{qi2017pointnet++} if the input changes to a single frame, while for our method, the anchor-based convolution can also be used for a single frame. 
We also test our anchor-based convolution on a single frame and compared it with PointNet++~\cite{qi2017pointnet++} in Table~\ref{table:1} and Table~\ref{table:2}. Ours obtains better performance than PointNet++~\cite{qi2017pointnet++}, which shows the superiority of structured feature learning by 3D convolution.

\subsubsection{Spatio-Temporal Attentive Embedding}
Introduced in Section~\ref{section:2}, an attentive embedding method is proposed during the feature embedding in the anchors. To demonstrate the effectiveness of this proposed method, we evaluate the networks with and without the attention both in action recognition and semantic segmentation. The experiment results of the two tasks are respectively shown in Table \ref{table:3} and Table \ref{table:4}. The attention method improves the accuracy of both tasks for various input numbers of frames. With the spatio-temporal attentive embedding, features are gathered in an adaptive weighted method, which achieves reasonable distribution of attention without loss of information.

\section{Conclusion}\label{5}

This paper proposes a novel anchor-based spatio-temporal attention 3D convolution (ASTA3DConv) and two networks based on ASTA3DConv to directly consume irregular dynamic 3D point cloud sequences. The anchor-based 3D convolution naturally aggregates structured information from unstructured point clouds. To adaptively fuse the spatio-temporal information from point cloud sequences, spatio-temporal attentive embedding is proposed and applied in the feature generation of anchors. Experiments on action recognition and semantic segmentation demonstrated the state-of-the-art performance of the proposed approach. Both action recognition and semantic segmentation can be regarded as the intelligent measurement and perception of the physical world, and are used in many high-level tasks. For example, after action recognition is performed through depth camera sensors, smart homes \cite{rafferty2017activity} can be realized through a wireless sensor network \cite{bassoli2017iot} based on the Internet of Things (IoT). The outdoor semantic scene understanding through the LiDAR measurement can be used in the localization \cite{yu2018ds}, depth estimation \cite{klingner2020self}, object tracking \cite{li2018stereo}, and navigation \cite{kostavelis2013learning} of robots.

\ifCLASSOPTIONcaptionsoff
\newpage
\fi



%
\bibliographystyle{IEEEtran}  
\bibliography{IEEEabrv, latestBIB}

\begin{thebibliography}{10}
\providecommand{\url}[1]{#1}
\csname url@samestyle\endcsname
\providecommand{\newblock}{\relax}
\providecommand{\bibinfo}[2]{#2}
\providecommand{\BIBentrySTDinterwordspacing}{\spaceskip=0pt\relax}
\providecommand{\BIBentryALTinterwordstretchfactor}{4}
\providecommand{\BIBentryALTinterwordspacing}{\spaceskip=\fontdimen2\font plus
\BIBentryALTinterwordstretchfactor\fontdimen3\font minus
  \fontdimen4\font\relax}
\providecommand{\BIBforeignlanguage}[2]{{%
\expandafter\ifx\csname l@#1\endcsname\relax
\typeout{** WARNING: IEEEtran.bst: No hyphenation pattern has been}%
\typeout{** loaded for the language `#1'. Using the pattern for}%
\typeout{** the default language instead.}%
\else
\language=\csname l@#1\endcsname
\fi
#2}}
\providecommand{\BIBdecl}{\relax}
\BIBdecl

\bibitem{zhang2019knowledge}
Y.~Zhang, G.~Tian, S.~Zhang, and C.~Li, ``A knowledge-based approach for
  multiagent collaboration in smart home: From activity recognition to guidance
  service,'' \emph{IEEE Trans. Instrum. Meas.}, vol.~69, no.~2, pp. 317--329,
  2020.

\bibitem{chen2019smartphone}
Z.~Chen, C.~Jiang, S.~Xiang, J.~Ding, M.~Wu, and X.~Li, ``Smartphone
  sensor-based human activity recognition using feature fusion and maximum full
  a posteriori,'' \emph{IEEE Trans. Instrum. Meas.}, vol.~69, no.~7, pp.
  3992--4001, 2019.

\bibitem{qiu2018rgb}
Z.~Qiu, Y.~Zhuang, F.~Yan, H.~Hu, and W.~Wang, ``Rgb-di images and full
  convolution neural network-based outdoor scene understanding for mobile
  robots,'' \emph{IEEE Trans. Instrum. Meas.}, vol.~68, no.~1, pp. 27--37,
  2018.

\bibitem{qi2017pointnet}
C.~R. Qi, H.~Su, K.~Mo, and L.~J. Guibas, ``Pointnet: Deep learning on point
  sets for 3d classification and segmentation,'' in \emph{Proc. CVPR}, 2017,
  pp. 652--660.

\bibitem{qi2017pointnet++}
C.~R. Qi, L.~Yi, H.~Su, and L.~J. Guibas, ``Pointnet++: Deep hierarchical
  feature learning on point sets in a metric space,'' in \emph{Proc. NeurIPS},
  2017, pp. 5099--5108.

\bibitem{wang2017cnn}
P.-S. Wang, Y.~Liu, Y.-X. Guo, C.-Y. Sun, and X.~Tong, ``O-cnn: Octree-based
  convolutional neural networks for 3d shape analysis,'' \emph{ACM Trans.
  Graph.}, vol.~36, no.~4, pp. 1--11, 2017.

\bibitem{riegler2017octnet}
G.~Riegler, A.~Osman~Ulusoy, and A.~Geiger, ``Octnet: Learning deep 3d
  representations at high resolutions,'' in \emph{Proc. CVPR}, 2017, pp.
  3577--3586.

\bibitem{graham20183d}
B.~Graham, M.~Engelcke, and L.~van~der Maaten, ``3d semantic segmentation with
  submanifold sparse convolutional networks,'' in \emph{Proc. CVPR}, 2018, pp.
  9224--9232.

\bibitem{kuang2019effective}
Z.~Kuang, J.~Yu, S.~Zhu, Z.~Li, and J.~Fan, ``Effective 3-d shape retrieval by
  integrating traditional descriptors and pointwise convolution,'' \emph{IEEE
  Trans. Multimedia}, vol.~21, no.~12, pp. 3164--3177, 2019.

\bibitem{mao2019interpolated}
J.~Mao, X.~Wang, and H.~Li, ``Interpolated convolutional networks for 3d point
  cloud understanding,'' in \emph{Proc. ICCV}, 2019, pp. 1578--1587.

\bibitem{hu2019randla}
Q.~Hu, B.~Yang, L.~Xie, S.~Rosa, Y.~Guo, Z.~Wang, N.~Trigoni, and A.~Markham,
  ``Randla-net: Efficient semantic segmentation of large-scale point clouds,''
  \emph{arXiv preprint arXiv:1911.11236}, 2019.

\bibitem{wang2020spherical}
G.~Wang, Y.~Yang, H.~Zhang, Z.~Liu, and H.~Wang, ``Spherical interpolated
  convolutional network with distance-feature density for 3d semantic
  segmentation of point clouds,'' \emph{arXiv preprint arXiv:2011.13784}, 2020.

\bibitem{choy20194d}
C.~Choy, J.~Gwak, and S.~Savarese, ``4d spatio-temporal convnets: Minkowski
  convolutional neural networks,'' in \emph{Proc. CVPR}, 2019, pp. 3075--3084.

\bibitem{luo2018fast}
W.~Luo, B.~Yang, and R.~Urtasun, ``Fast and furious: Real time end-to-end 3d
  detection, tracking and motion forecasting with a single convolutional net,''
  in \emph{Proc. CVPR}, 2018, pp. 3569--3577.

\bibitem{liu2019meteornet}
X.~Liu, M.~Yan, and J.~Bohg, ``Meteornet: Deep learning on dynamic 3d point
  cloud sequences,'' in \emph{Proc. ICCV}, 2019, pp. 9246--9255.

\bibitem{wang2020hierarchical}
G.~Wang, X.~Wu, Z.~Liu, and H.~Wang, ``Hierarchical attention learning of scene
  flow in 3d point clouds,'' \emph{IEEE Trans. Image Process.}, vol.~30, pp.
  5168--5181, 2021.

\bibitem{zhang2019multi}
W.~Zhang, X.~He, X.~Yu, W.~Lu, Z.~Zha, and Q.~Tian, ``A multi-scale
  spatial-temporal attention model for person re-identification in videos,''
  \emph{IEEE Trans. Image Process.}, vol.~29, pp. 3365--3373, 2019.

\bibitem{li2018unified}
D.~Li, T.~Yao, L.-Y. Duan, T.~Mei, and Y.~Rui, ``Unified spatio-temporal
  attention networks for action recognition in videos,'' \emph{IEEE Trans.
  Multimedia}, vol.~21, no.~2, pp. 416--428, 2018.

\bibitem{zhong2013video}
S.-h. Zhong, Y.~Liu, F.~Ren, J.~Zhang, and T.~Ren, ``Video saliency detection
  via dynamic consistent spatio-temporal attention modelling,'' in \emph{Proc.
  AAAI}, 2013.

\bibitem{li2010action}
W.~Li, Z.~Zhang, and Z.~Liu, ``Action recognition based on a bag of 3d
  points,'' in \emph{Proc. CVPRW}.\hskip 1em plus 0.5em minus 0.4em\relax IEEE,
  2010, pp. 9--14.

\bibitem{ros2016synthia}
G.~Ros, L.~Sellart, J.~Materzynska, D.~Vazquez, and A.~M. Lopez, ``The synthia
  dataset: A large collection of synthetic images for semantic segmentation of
  urban scenes,'' in \emph{Proc. CVPR}, 2016, pp. 3234--3243.

\bibitem{su2018splatnet}
H.~Su, V.~Jampani, D.~Sun, S.~Maji, E.~Kalogerakis, M.-H. Yang, and J.~Kautz,
  ``Splatnet: Sparse lattice networks for point cloud processing,'' in
  \emph{Proc. CVPR}, 2018, pp. 2530--2539.

\bibitem{xu2018spidercnn}
Y.~Xu, T.~Fan, M.~Xu, L.~Zeng, and Y.~Qiao, ``Spidercnn: Deep learning on point
  sets with parameterized convolutional filters,'' in \emph{Proc. ECCV}, 2018,
  pp. 87--102.

\bibitem{komarichev2019cnn}
A.~Komarichev, Z.~Zhong, and J.~Hua, ``A-cnn: Annularly convolutional neural
  networks on point clouds,'' in \emph{Proc. CVPR}, 2019, pp. 7421--7430.

\bibitem{lei2019octree}
H.~Lei, N.~Akhtar, and A.~Mian, ``Octree guided cnn with spherical kernels for
  3d point clouds,'' in \emph{Proc. CVPR}, 2019, pp. 9631--9640.

\bibitem{zhao2019pointweb}
H.~Zhao, L.~Jiang, C.-W. Fu, and J.~Jia, ``Pointweb: Enhancing local
  neighborhood features for point cloud processing,'' in \emph{Proc. CVPR},
  2019, pp. 5565--5573.

\bibitem{wang2019dynamic}
Y.~Wang, Y.~Sun, Z.~Liu, S.~E. Sarma, M.~M. Bronstein, and J.~M. Solomon,
  ``Dynamic graph cnn for learning on point clouds,'' \emph{ACM Trans. Graph.},
  vol.~38, no.~5, pp. 1--12, 2019.

\bibitem{zhang2019shellnet}
Z.~Zhang, B.-S. Hua, and S.-K. Yeung, ``Shellnet: Efficient point cloud
  convolutional neural networks using concentric shells statistics,'' in
  \emph{Proc. ICCV}, 2019, pp. 1607--1616.

\bibitem{liu2019pvcnn}
Z.~Liu, H.~Tang, Y.~Lin, and S.~Han, ``Point-voxel cnn for efficient 3d deep
  learning,'' in \emph{Proc. NeurIPS}, 2019.

\bibitem{xie2020grnet}
H.~Xie, H.~Yao, S.~Zhou, J.~Mao, S.~Zhang, and W.~Sun, ``Grnet: gridding
  residual network for dense point cloud completion,'' in \emph{Proc.
  ECCV}.\hskip 1em plus 0.5em minus 0.4em\relax Springer, 2020, pp. 365--381.

\bibitem{thomas2019kpconv}
H.~Thomas, C.~R. Qi, J.-E. Deschaud, B.~Marcotegui, F.~Goulette, and L.~J.
  Guibas, ``Kpconv: Flexible and deformable convolution for point clouds,'' in
  \emph{Proc. ICCV}, 2019, pp. 6411--6420.

\bibitem{wu2020pointpwc}
W.~Wu, Z.~Y. Wang, Z.~Li, W.~Liu, and L.~Fuxin, ``Pointpwc-net: Cost volume on
  point clouds for (self-) supervised scene flow estimation,'' in \emph{Proc.
  ECCV}.\hskip 1em plus 0.5em minus 0.4em\relax Springer, 2020, pp. 88--107.

\bibitem{gu2019hplflownet}
X.~Gu, Y.~Wang, C.~Wu, Y.~J. Lee, and P.~Wang, ``Hplflownet: Hierarchical
  permutohedral lattice flownet for scene flow estimation on large-scale point
  clouds,'' in \emph{Proc. CVPR}, 2019, pp. 3254--3263.

\bibitem{liu2019flownet3d}
X.~Liu, C.~R. Qi, and L.~J. Guibas, ``Flownet3d: Learning scene flow in 3d
  point clouds,'' in \emph{Proc. CVPR}, 2019, pp. 529--537.

\bibitem{min2020efficient}
Y.~Min, Y.~Zhang, X.~Chai, and X.~Chen, ``An efficient pointlstm for point
  clouds based gesture recognition,'' in \emph{Proc. CVPR}, 2020, pp.
  5761--5770.

\bibitem{Donahue_2015_CVPR}
J.~Donahue, L.~Anne~Hendricks, S.~Guadarrama, M.~Rohrbach, S.~Venugopalan,
  K.~Saenko, and T.~Darrell, ``Long-term recurrent convolutional networks for
  visual recognition and description,'' in \emph{Proc. CVPR}, June 2015.

\bibitem{hochreiter1997long}
S.~Hochreiter and J.~Schmidhuber, ``Long short-term memory,'' \emph{Neural
  Comput.}, vol.~9, no.~8, pp. 1735--1780, 1997.

\bibitem{caoasap}
H.~Cao, Y.~Lu, C.~Lu, B.~Pang, G.~Liu, and A.~Yuille, ``Asap-net: Attention and
  structure aware point cloud sequence segmentation,'' in \emph{Proc. BMVC},
  2020.

\bibitem{fan2021point}
H.~Fan, Y.~Yang, and M.~Kankanhalli, ``Point 4d transformer networks for
  spatio-temporal modeling in point cloud videos,'' in \emph{Proc. CVPR}, 2021,
  pp. 14\,204--14\,213.

\bibitem{hales1998overview}
T.~C. Hales, ``An overview of the kepler conjecture,'' \emph{arXiv preprint
  math/9811071}, 1998.

\bibitem{szpiro2003mathematics}
G.~Szpiro, ``Mathematics: Does the proof stack up?'' 2003.

\bibitem{hales2017formal}
T.~Hales, M.~Adams, G.~Bauer, T.~D. Dang, J.~Harrison, H.~Le~Truong,
  C.~Kaliszyk, V.~Magron, S.~McLaughlin, T.~T. Nguyen \emph{et~al.}, ``A formal
  proof of the kepler conjecture,'' in \emph{Forum Math. Pi}, vol.~5.\hskip 1em
  plus 0.5em minus 0.4em\relax Cambridge University Press, 2017.

\bibitem{ronneberger2015u}
O.~Ronneberger, P.~Fischer, and T.~Brox, ``U-net: Convolutional networks for
  biomedical image segmentation,'' in \emph{Proc. MICCAI}.\hskip 1em plus 0.5em
  minus 0.4em\relax Springer, 2015, pp. 234--241.

\bibitem{oreifej2013hon4d}
O.~Oreifej and Z.~Liu, ``Hon4d: Histogram of oriented 4d normals for activity
  recognition from depth sequences,'' in \emph{Proc. CVPR}, 2013, pp. 716--723.

\bibitem{wang2012mining}
J.~Wang, Z.~Liu, Y.~Wu, and J.~Yuan, ``Mining actionlet ensemble for action
  recognition with depth cameras,'' in \emph{Proc. CVPR}.\hskip 1em plus 0.5em
  minus 0.4em\relax IEEE, 2012, pp. 1290--1297.

\bibitem{presti2014gesture}
L.~L. Presti, M.~La~Cascia, S.~Sclaroff, and O.~Camps, ``Gesture modeling by
  hanklet-based hidden markov model,'' in \emph{Proc. ACCV}.\hskip 1em plus
  0.5em minus 0.4em\relax Springer, 2014, pp. 529--546.

\bibitem{vemulapalli2014human}
R.~Vemulapalli, F.~Arrate, and R.~Chellappa, ``Human action recognition by
  representing 3d skeletons as points in a lie group,'' in \emph{Proc. CVPR},
  2014, pp. 588--595.

\bibitem{devanne20143}
M.~Devanne, H.~Wannous, S.~Berretti, P.~Pala, M.~Daoudi, and A.~Del~Bimbo,
  ``3-d human action recognition by shape analysis of motion trajectories on
  riemannian manifold,'' \emph{IEEE Trans. Cybern.}, vol.~45, no.~7, pp.
  1340--1352, 2014.

\bibitem{song2014body}
Y.~Song, J.~Tang, F.~Liu, and S.~Yan, ``Body surface context: A new robust
  feature for action recognition from depth videos,'' \emph{IEEE Trans.
  Circuits Syst. Video Technol.}, vol.~24, no.~6, pp. 952--964, 2014.

\bibitem{ioffe2015batch}
S.~Ioffe and C.~Szegedy, ``Batch normalization: Accelerating deep network
  training by reducing internal covariate shift,'' \emph{arXiv preprint
  arXiv:1502.03167}, 2015.

\bibitem{kingma2014adam}
D.~P. Kingma and J.~Ba, ``Adam: A method for stochastic optimization,''
  \emph{arXiv preprint arXiv:1412.6980}, 2014.

\bibitem{rafferty2017activity}
J.~Rafferty, C.~D. Nugent, J.~Liu, and L.~Chen, ``From activity recognition to
  intention recognition for assisted living within smart homes,'' \emph{IEEE
  Trans. Human-Mach. Syst.}, vol.~47, no.~3, pp. 368--379, 2017.

\bibitem{bassoli2017iot}
M.~Bassoli, V.~Bianchi, I.~De~Munari, and P.~Ciampolini, ``An iot approach for
  an aal wi-fi-based monitoring system,'' \emph{IEEE Trans. Instrum. Meas.},
  vol.~66, no.~12, pp. 3200--3209, 2017.

\bibitem{yu2018ds}
C.~Yu, Z.~Liu, X.-J. Liu, F.~Xie, Y.~Yang, Q.~Wei, and Q.~Fei, ``Ds-slam: A
  semantic visual slam towards dynamic environments,'' in \emph{Proc.
  IROS}.\hskip 1em plus 0.5em minus 0.4em\relax IEEE, 2018, pp. 1168--1174.

\bibitem{klingner2020self}
M.~Klingner, J.-A. Term{\"o}hlen, J.~Mikolajczyk, and T.~Fingscheidt,
  ``Self-supervised monocular depth estimation: Solving the dynamic object
  problem by semantic guidance,'' in \emph{Proc. ECCV}.\hskip 1em plus 0.5em
  minus 0.4em\relax Springer, 2020, pp. 582--600.

\bibitem{li2018stereo}
P.~Li, T.~Qin \emph{et~al.}, ``Stereo vision-based semantic 3d object and
  ego-motion tracking for autonomous driving,'' in \emph{Proc. ECCV}, 2018, pp.
  646--661.

\bibitem{kostavelis2013learning}
I.~Kostavelis and A.~Gasteratos, ``Learning spatially semantic representations
  for cognitive robot navigation,'' \emph{Rob Auton Syst}, vol.~61, no.~12, pp.
  1460--1475, 2013.

\end{thebibliography}

%
\vspace{-1.0cm}
\begin{IEEEbiography}[{\includegraphics[width=1in,height=1.25in,clip,keepaspectratio]{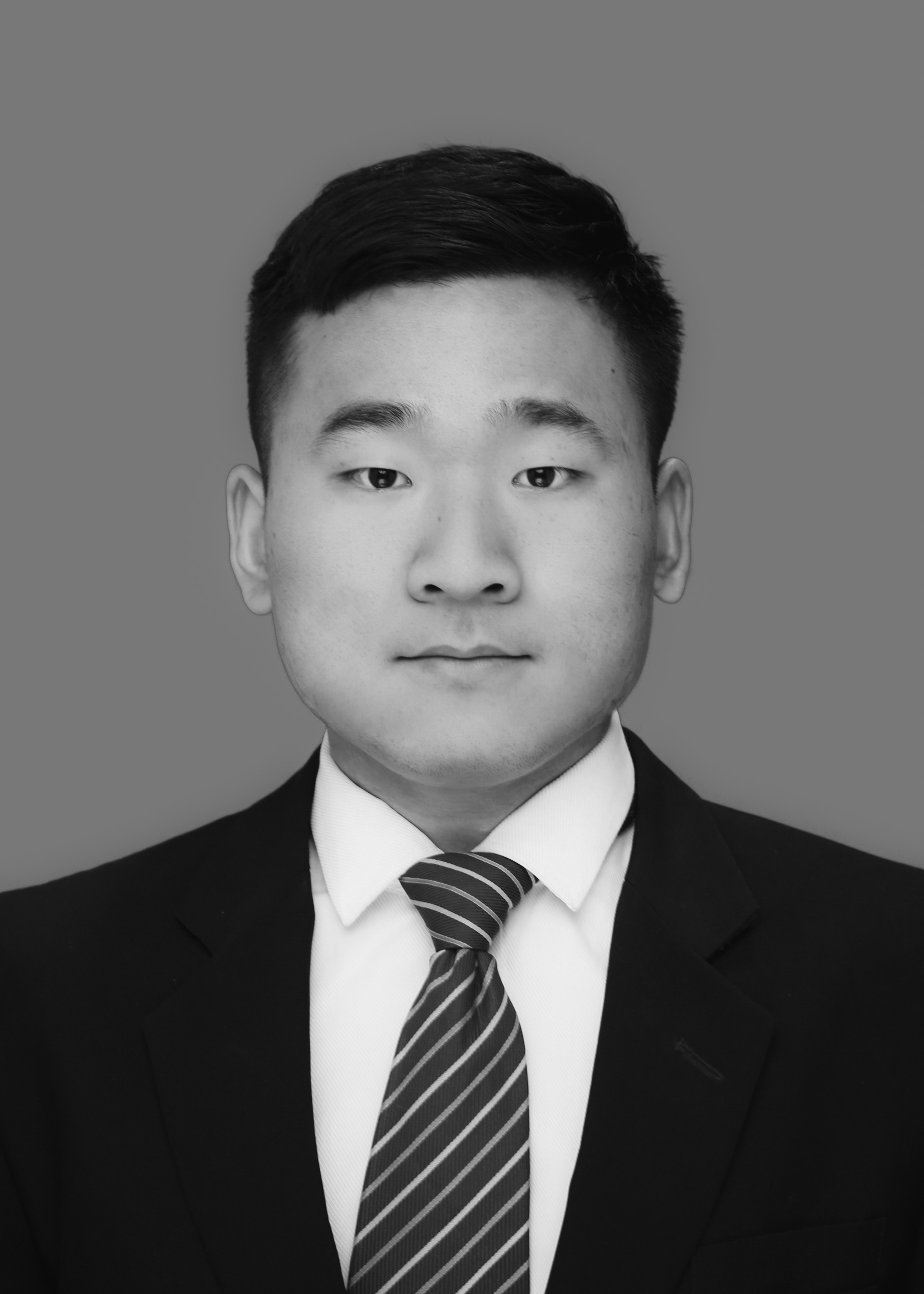}}]{Guangming Wang} received the B.S. degree from Department of Automation from Central South University, Changsha, China, in 2018. He is currently pursuing the Ph.D. degree in Control Science and Engineering with Shanghai Jiao Tong University. His current research interests include SLAM and computer vision, in particular, deep learning on point clouds.
\end{IEEEbiography}
\begin{IEEEbiography}[{\includegraphics[width=1in,height=1.25in,clip,keepaspectratio]{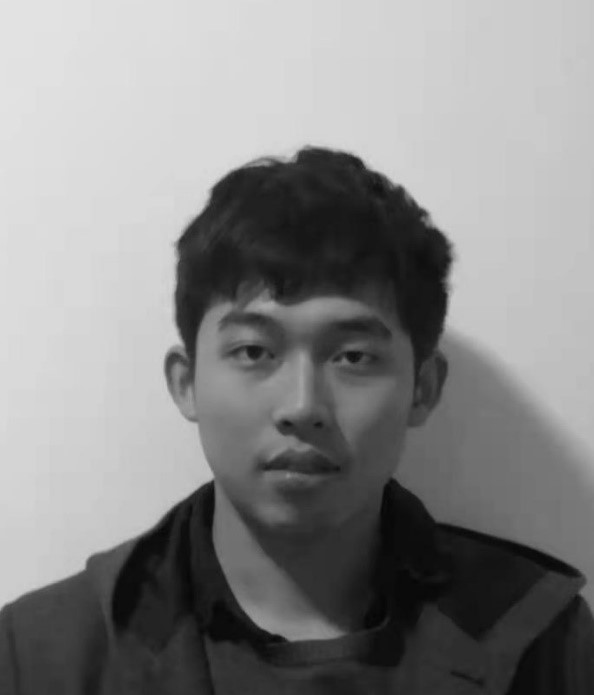}}]{Muyao Chen} 
	is currently pursuing the B.S. degree in Department of Computer Science and Engineering, Shanghai Jiao Tong University. His latest research interests include 3D point clouds and computer vision.
\end{IEEEbiography}
\vspace{-1.0cm}

\begin{IEEEbiography}[{\includegraphics[width=1in,height=1.25in,clip,keepaspectratio]{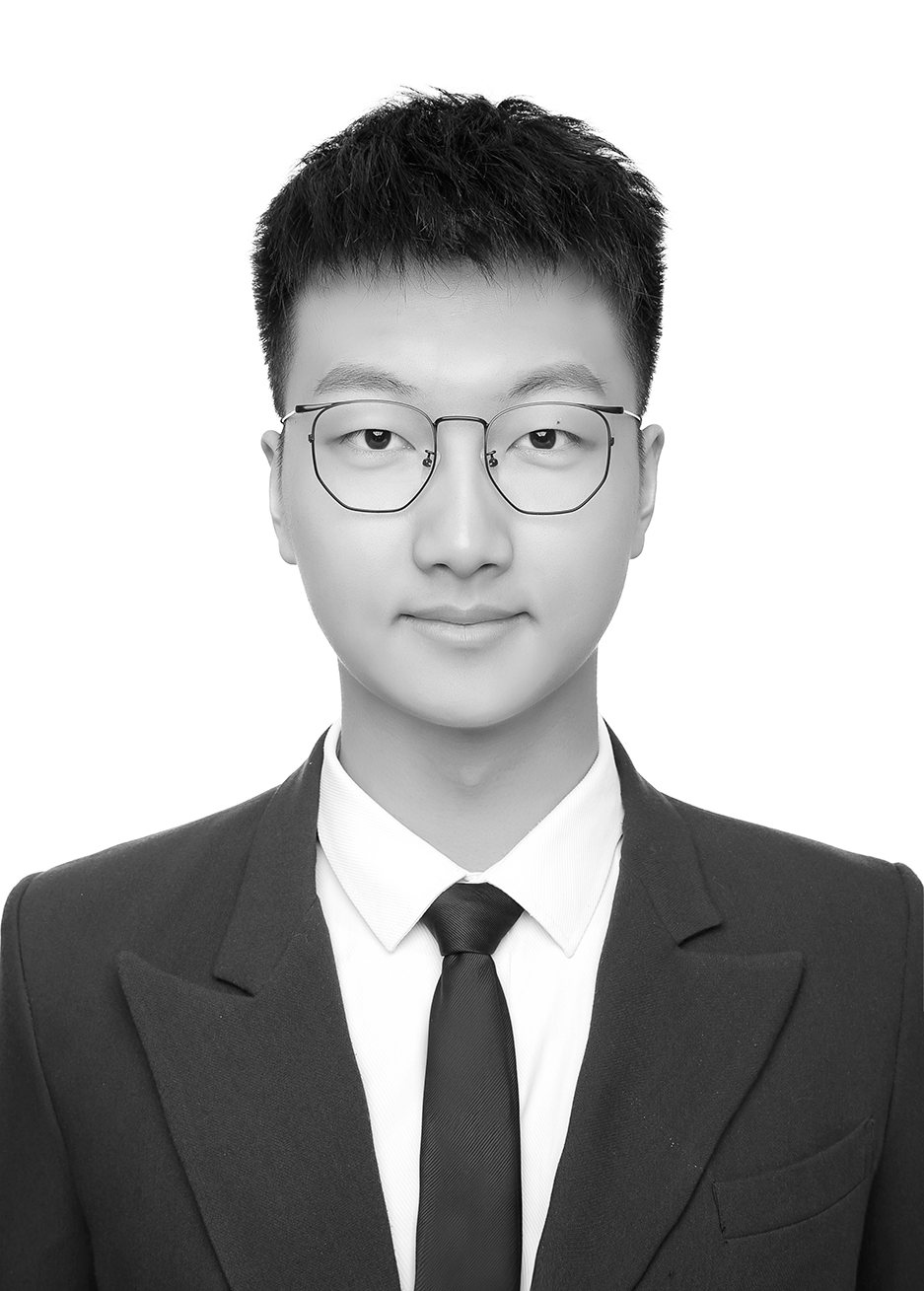}}]{Hanwen Liu} 
	is currently pursuing the B.Eng. degree in Department of Computer Science and Engineering, Shanghai Jiao Tong University. His latest research interests include 3D point clouds and computer vision.
\end{IEEEbiography}
\vspace{-1.0cm}
\begin{IEEEbiography}[{\includegraphics[width=1in,height=1.25in,clip,keepaspectratio]{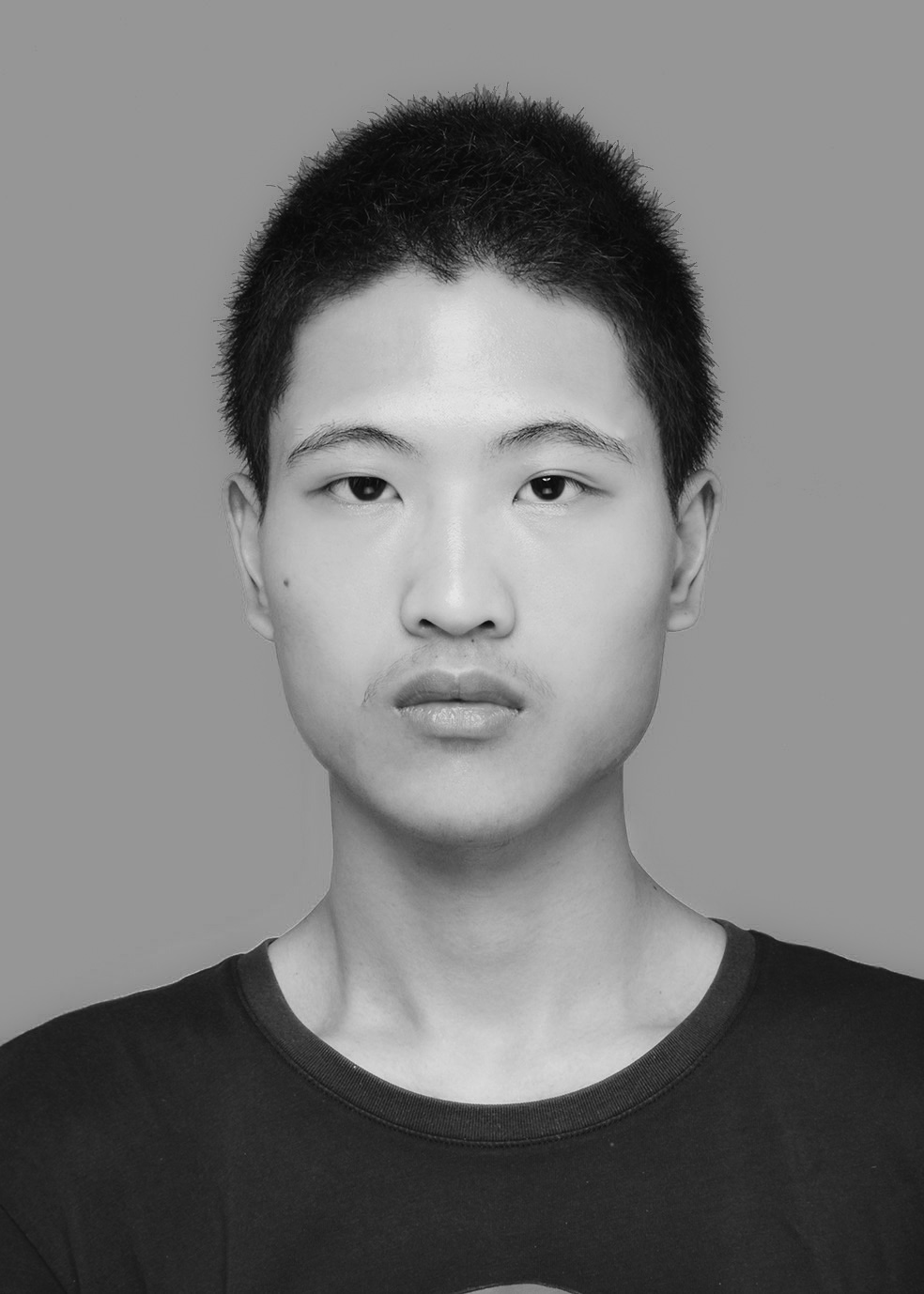}}]{Yehui Yang} 
	is currently pursuing the B.S. degree in Department of Automation, Shanghai Jiao Tong University. His latest research interests include SLAM and computer vision.
\end{IEEEbiography}
\vspace{-1.0cm}

\begin{IEEEbiography}[{\includegraphics[width=1in,height=1.25in,clip,keepaspectratio]{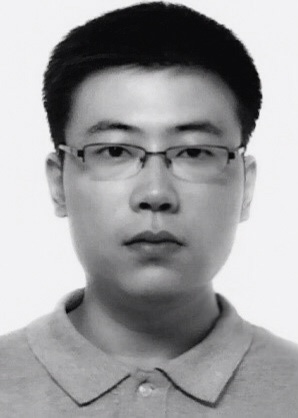}}]{Zhe Liu} received his B.S. degree in Automation from Tianjin University, Tianjin, China, in 2010, and Ph.D. degree in Control Technology and Control Engineering from Shanghai Jiao Tong University, Shanghai, China, in 2016. From 2017 to 2020, he was a Post-Doctoral Fellow with the Department of Mechanical and Automation Engineering, The Chinese University of Hong Kong, Hong Kong. He is currently a Research Associate with the Department of Computer Science and Technology, University of Cambridge. His research interests include autonomous mobile robot, multirobot cooperation and autonomous driving system.
\end{IEEEbiography}
\vspace{-1.0cm}

\begin{IEEEbiography}[{\includegraphics[width=1in,height=1.25in,clip,keepaspectratio]{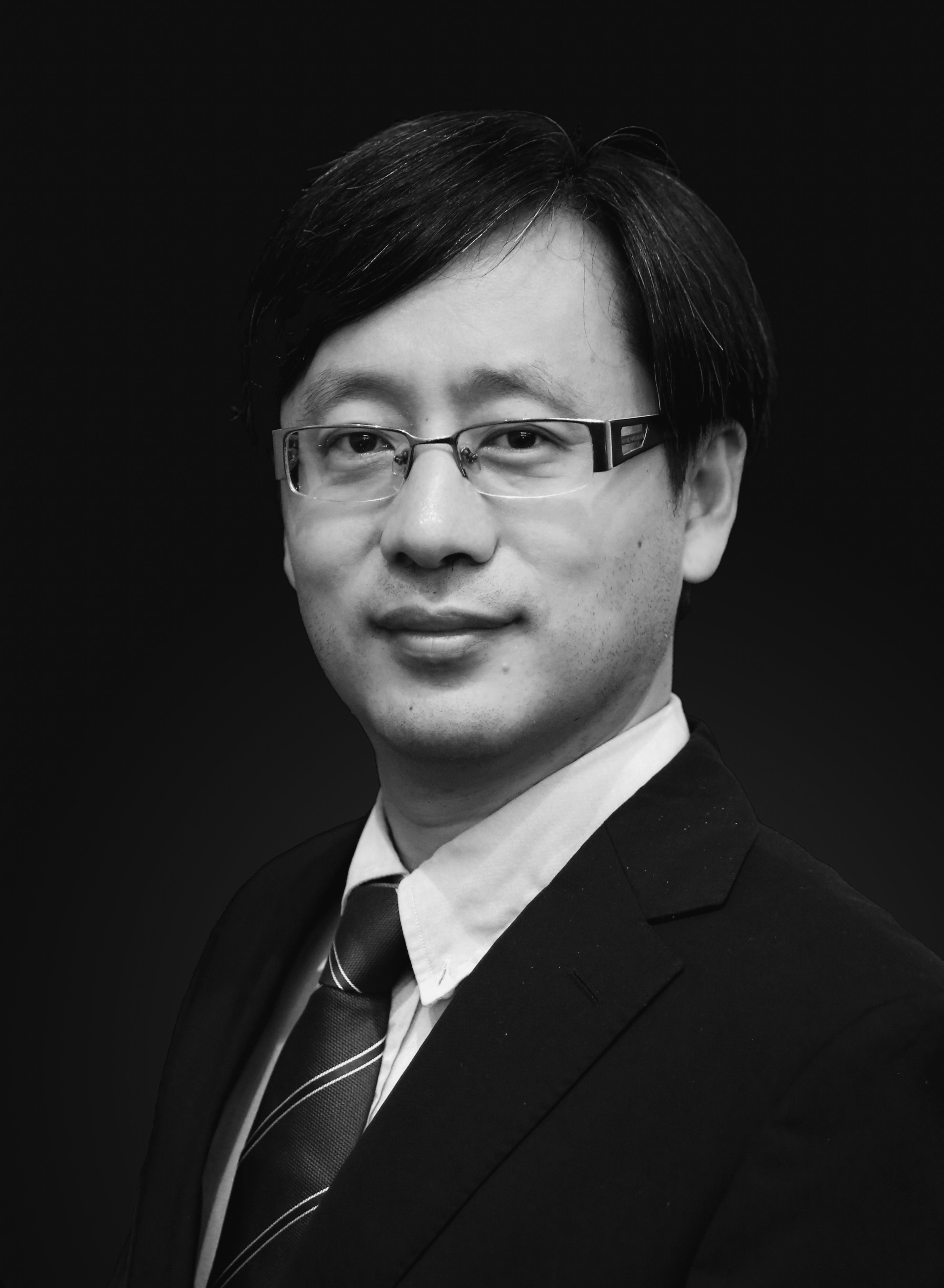}}]{Hesheng Wang} 
	(SM'15) received the B.Eng. degree in electrical engineering from the Harbin Institute of Technology, Harbin, China, in 2002, and the M.Phil. and Ph.D. degrees in automation and computer-aided engineering from The Chinese University of Hong Kong, Hong Kong, in 2004 and 2007, respectively. He was a Post-Doctoral Fellow and Research Assistant with the Department of Mechanical and Automation Engineering, The Chinese University of Hong Kong, from 2007 to 2009. He is currently a Professor with the Department of Automation, Shanghai Jiao Tong University, Shanghai, China. His current research interests include visual servoing, service robot, adaptive robot control, and autonomous driving. 
	Dr. Wang is an Associate Editor of Assembly Automation and the International Journal of Humanoid Robotics, a Technical Editor of the IEEE/ASME TRANSACTIONS ON MECHATRONICS. He served as an Associate Editor of the IEEE TRANSACTIONS ON ROBOTICS from 2015 to 2019. He was the General Chair of the IEEE RCAR 2016, and the Program Chair of the IEEE ROBIO 2014 and IEEE/ASME AIM 2019.	
\end{IEEEbiography}




\end{document}